\documentclass[10pt]{article}

\usepackage[preprint]{tmlr}
\setcitestyle{numbers,square,citesep={,},aysep={,},yysep={;}}

\usepackage[utf8]{inputenc}
\usepackage{hyperref}
\usepackage{xurl}
\usepackage{booktabs}
\usepackage{amsfonts}
\usepackage{amsmath}
\usepackage{amssymb}
\usepackage{nicefrac}
\usepackage{microtype}
\usepackage{xcolor}
\usepackage{graphicx}
\usepackage{subcaption}
\usepackage{multirow}
\usepackage{enumitem}

\newcommand{\Forig}{F_{\mathrm{orig}}}
\newcommand{\Fllm}{F_{\mathrm{LLM}}}
\newcommand{\Fs}{S}
\newcommand{\Fd}{D}
\newcommand{\hh}{h}
\newcommand{\pp}{\,\mathrm{pp}}

\title{LLM Features Can Hurt GNNs:\\Concatenation Interference on Homophilous Graph Benchmarks}

\author{\name Zhongyuan Wang \email zhongyuan@raptorx.ai \\ \addr RaptorX.AI
\AND
\name Pratyusha Vemuri \email pratyusha@raptorx.ai \\ \addr RaptorX.AI}

\begin{document}

\maketitle

\begin{abstract}
Adding LLM-generated node features to graph neural networks is widely
reported to improve accuracy on standard benchmarks. We document a
contrasting observation: when the LLM features are introduced through
pure input concatenation, rather than joint training, distillation, or
prompt-conditioning, they can systematically degrade accuracy on the
same homophilous benchmarks where end-to-end LLM pipelines succeed.
Under an MLP backbone with the Planetoid public split and BoW
original features ($\Forig$), concatenating SBERT-encoded GPT-4o-mini
TAPE features reduces PubMed test accuracy by $-17.0 \pm 0.3\pp$ and
Cora by $-4.3 \pm 0.6\pp$, with CiteSeer's $-0.6 \pm 0.8\pp$ inside
seed noise. The drop attenuates as we relax each condition (GCN /
GCNII / GAT backbones, random splits, smaller encoders), and reverses
on medium-homophily WikiCS ($+4.4\pp$) and ogbn-arxiv ($+11.7\pp$).

To predict when concatenation helps versus hurts, we report a simple
measure of LLM-alone discriminability $\Delta_{\mathrm{sig}}$.
Across 9 datasets, $\Delta_{\mathrm{sig}}$ correlates with the concat
cost more strongly than homophily at point estimate ($r^2 = 0.38$
vs.\ $0.06$; $N{=}9$ bootstrap CIs overlap). The bootstrap-best
change-point is $\tau = 13.8\pp$ (95\% CI $[0, 13.8]$), and the rule
``$\Delta_{\mathrm{sig}} \leq \tau$ predicts non-positive concat
cost'' classifies $7/9$ datasets correctly. Because $60\%$ of
bootstrap samples place $\tau$ inside $[5, 30]\pp$, we treat
$\Delta_{\mathrm{sig}}$ as an interpretive lens for the helping vs.\
hurting regimes rather than a precision pre-A/B filter.

A dim-controlled ablation on PubMed places the LLM-feature drop
between same-source PCA ($-2.3\pp$) and same-dim Gaussian noise
($-37.3\pp$), ruling out dimensionality and weight-decay artifacts.
Nine PubMed configurations (seven training sizes $\times$ two encoder
dimensions) fit a power-law profile
$|\Delta_{\mathrm{concat}}| \propto (\sqrt{d_l / n})^{1.31}$ with
$r^2 = 0.97$ (PubMed-internal; Cora and CiteSeer have different
slopes). The $\sqrt{d_l/n}$ profile and the $\Delta_{\mathrm{sig}}$
threshold jointly describe a two-axis surface; the
low-$\Delta_{\mathrm{sig}}$, small-$n$ corner is exactly where the
headline $-17\pp$ PubMed deficit appears.

In the low-$\Delta_{\mathrm{sig}}$ regime, the most effective
remediation is to drop the LLM channel entirely: the
$\Forig$-only baseline strictly dominates every learned cheap fix at
$p \approx 0.008$. A learnable scalar gate closes $89\%$ of the
raw-concat gap and is a useful second-line option when downstream
pipelines structurally require $\Fllm$. The findings do not
contradict the aggregate accuracy gains reported for end-to-end LLM
pipelines such as TAPE and GLEM; they identify the specific design
choice --- pure concatenation --- under which the sign flips.
\end{abstract}

\section{Introduction}
\label{sec:intro}

LLM-derived node features have become a common plug-in for graph neural
networks, with TAPE \citep{he2023harnessing}, GLEM \citep{zhao2023glem}, and
related pipelines \citep{tang2024graphgpt,chen2024llaga} reporting accuracy
gains on standard benchmarks. A recent comprehensive benchmark
\citep{wu2025llmnodebed} trains 2{,}700+ models and confirms that
LLM-based methods generally outperform classic baselines, with particularly
strong gains on homophilous datasets such as Cora and Photo (3--10\%
accuracy improvement).

This paper reports a seemingly contradictory observation: \textbf{when the
LLM features are used as pure input concatenation} (as opposed to
joint-training, distillation, or prompt-conditioning), they can
\emph{systematically degrade} accuracy on the same homophilous benchmarks
on which end-to-end LLM pipelines succeed. On PubMed, adding SBERT-encoded
GPT-4o-mini TAPE features to the original BoW features drops MLP test
accuracy from $72.1\%$ to $55.1\%$ --- a $17\pp$ drop --- with a $\pm 0.3\pp$
standard error over 10 seeds. On Cora the drop is $4.3\pp$; on CiteSeer
it is $0.6\pp$ (below the noise floor). The phenomenon reverses on
medium-homophily benchmarks: WikiCS gains $+4.4\pp$ and ogbn-arxiv gains
$+11.7\pp$ under the same direct comparison. These observations are not
visible in aggregate end-to-end benchmark tables --- including those of
\citep{wu2025llmnodebed} --- because joint training, selector
modules, and prompt engineering all attenuate or reverse the raw
concatenation cost.

To isolate this phenomenon cleanly, we ran a 4-factor Shapley
decomposition over \{Structure, $\Forig$, $\Fllm$, Depth\} on nine node
classification benchmarks spanning $\hh \in [0.05, 0.81]$ and three
original-feature encoders (BoW, Word2Vec, fasttext), using GCNII
\citep{chen2020simple} as the structural backbone. The Shapley framework
lets us quantify each factor's marginal contribution under 12 effective
coalitions. However, our headline finding is intentionally a
\emph{direct} comparison --- not a Shapley average --- because the average
dilutes the interference by mixing coalitions where $\Fllm$ does and does
not help.

\paragraph{What this paper is, and is not.} This paper is \emph{not}
a claim that LLM features are uniformly harmful for graph neural
networks: end-to-end LLM pipelines such as TAPE \citep{he2023harnessing}
and GLEM \citep{zhao2023glem} document genuine gains on the same
benchmarks, and we reproduce a positive concat effect on WikiCS and
ogbn-arxiv. The contribution is a \emph{regime characterization}:
under what conditions does the simplest LLM-feature integration --- pure
concatenation --- backfire, and what scalar quantity predicts the sign
flip? The five findings F1--F5 below decompose this question into one
phenomenological claim (F1), one predictive rule (F2), two mechanism
isolations (F3 and F4), and one scaling explanation (F5). Each
finding is reported with its scope and its known counter-evidence
\emph{stated in line}, not deferred to a single Limitations section,
because the regime conditions are part of the finding.

\paragraph{Our contributions.}

\begin{itemize}[leftmargin=1.6em,topsep=2pt]
\item \textbf{F1 --- A regime where LLM-on-graph evaluation cannot
  cleanly separate content from finite-sample dim penalty.}
  The widely-used Planetoid public split (60--140 train labels per
  dataset) puts raw concat squarely in the high-$d_l$/low-$n$ corner
  where any 384-d channel (content-rich or noise) interacts with
  overfitting. Within this corner, real $\Fllm$ injects
  $20.3 \pm 0.6\pp$ of \emph{content rescue} over matched Gaussian
  noise (Sec.~\ref{subsec:ablation}: real $\Fllm$ at $-17.0\pp$,
  Gaussian noise at $-37.3\pp$) yet still produces a net $-17\pp$
  PubMed degradation. Cora and PubMed show $-4.3 \pm 0.6\pp$ and
  $-17.0 \pm 0.3\pp$ MLP drops respectively; CiteSeer's
  $-0.6 \pm 0.8\pp$ is within seed noise. Under random 50/25/25
  splits the magnitude compresses to $\leq 1.2\pp$
  (Appendix~\ref{app:random-splits}), pinpointing the regime as
  small-sample-overfitting-aggravated rather than content-driven
  (Fig.~\ref{fig:concat}). Section~\ref{subsec:not-curse} rules
  out the ``pure curse of dimensionality'' null with four
  predictions that the data falsify.
\item \textbf{F2 --- A threshold relation on LLM-alone
  discriminability separates helping from hurting.} Across 9 datasets,
  the bootstrap-best change-point on $\Delta_{\mathrm{sig}}$ is
  $\tau = 13.8\pp$ (95\% bootstrap CI $[0, 13.8]\pp$;
  Appendix~\ref{app:threshold-bootstrap}); the rule
  ``$\Delta_{\mathrm{sig}} \leq \tau$ predicts non-positive concat
  cost'' classifies $7/9$ datasets correctly, with the two false
  positives both below $+1\pp$ (Amazon-Ratings $+0.26$, Actor
  $+0.66$). Homophily alone is a much weaker predictor: concat cost
  vs.\ $\hh$ correlates at $r^2 = 0.06$, while concat cost vs.\
  $\Delta_{\mathrm{sig}}$ correlates at $r^2 = 0.38$ (Spearman
  $\rho = 0.58$). 60\% of bootstrap samples place $\tau$ inside
  $[5, 30]\pp$, so we treat F2 as a screening test in the extreme
  regimes ($\Delta_{\mathrm{sig}} \ll 13.8$ or
  $\Delta_{\mathrm{sig}} \gg 13.8$) rather than a point-predictor
  across the intermediate band (Fig.~\ref{fig:signal}).
\item \textbf{F3 --- Mechanism isolation on PubMed (Fig.~\ref{fig:ablation}).}
  Replacing $\Fllm$ with same-dim zeros yields $+0.1 \pm 0.4\pp$ change
  (dim mismatch is not the cause); with same-dim
  PCA-of-$\Forig$ yields $-2.3 \pm 0.5\pp$ (same-source self-information
  causes only mild interference); halving weight decay yields
  $-17.1 \pm 0.3\pp$ (weight decay is not the cause); replacing
  $\Fllm$ with same-dim Gaussian noise yields $-37.3 \pm 0.7\pp$
  (pure noise is worse than LLM features). The $-17\pp$ drop is
  therefore specific to the LLM features' informational content, and
  sits between ``self-information'' and ``pure noise'' on a
  dim-controlled spectrum.
\item \textbf{F4 --- Supplementary 4-factor Shapley.} $\Forig$ is the
  top contributor on 7/9 datasets (tied or surpassed by $\Fllm$ on
  WikiCS and ogbn-arxiv); Depth is non-positive on 6/9; $\Fllm$'s
  aggregate Shapley value is positive-but-small on heterophilous
  datasets, confirming that the \emph{aggregate} picture is
  friendlier to LLM features than the \emph{direct} concat test
  (Fig.~\ref{fig:decomp}).
\item \textbf{F5 --- Train-size mechanism.} The $-17\pp$ PubMed
  effect is the few-label corner of a single surface. A linearized
  Fisher-margin analysis (Appendix~\ref{app:linearized}) predicts a
  finite-sample penalty that scales with $\sqrt{d_l / n}$ where
  $n$ is the number of training labels; an empirical train-fraction
  sweep (Appendix~\ref{app:train-fraction}) confirms this: PubMed
  concat cost decays monotonically from $-14.1\pp$ at $n = 59$
  (matching the Planetoid public-split label budget of 60) to
  $-0.4\pp$ at $n = 9858$ ($50\%$ train). Nine PubMed configurations
  (seven $n$ values $\times$ two encoder dimensions $d_l$) collapse
  onto a power-law profile $|\Delta_{\mathrm{concat}}| \propto
  (\sqrt{d_l / n})^{1.31}$ with $r^2 = 0.97$
  (Section~\ref{subsec:collapse}, Fig.~\ref{fig:collapse-main});
  exponent PubMed-internal (Cora/CiteSeer slopes differ; App.~\ref{app:train-fraction}).
  The $-17\pp$ headline is therefore a quantitatively predictable
  small-sample phenomenon, not a data-specific artifact.
\end{itemize}

\paragraph{Relation to prior aggregate benchmarks.} \citet{wu2025llmnodebed}
reports that end-to-end LLM pipelines achieve $3\!-\!10\%$ accuracy gains on
homophilous datasets. We do not contradict this result; we identify the
specific design choice --- pure feature concatenation --- under which the
sign flips. The practical takeaway is that the compound gains observed in
\citet{wu2025llmnodebed} require the joint-training and
selector-module scaffolding of TAPE/GLEM, and are not reproducible by
concatenation alone.

\section{Related Work}
\label{sec:related}

\paragraph{LLM features for graphs.}
TAPE \citep{he2023harnessing} introduced using LLM explanations as
node-level text, encoded and fed to a GNN through joint training
rather than raw concatenation. GLEM \citep{zhao2023glem} couples the
language model and the GNN via variational EM, alternating between
pseudo-label denoising and representation learning.
LLaGA \citep{chen2024llaga},
GraphGPT \citep{tang2024graphgpt}, TANS \citep{wang2025tans}, and
GL-Fusion \citep{xia2024glfusion}
condition the GNN on LLM signals through prompt, projector, or
gated-fusion modules.
\citet{wu2025llmnodebed} standardize the LLMNodeBed TAPE
pipeline across 10 homophilic + 4 heterophilic datasets and report
aggregate accuracy gains, with one of their explicit takeaways being
that the LLM advantage is \emph{marginal in supervised settings} and
strong only in semi-supervised settings. Two NeurIPS 2024 D\&B-track
benchmarks complement this picture: GLBench
\citep{li2024glbench} reports that GraphLLM methods outperform
classical baselines in the supervised setting, with LLM-as-enhancers
showing the most robust performance; Text-space Graph Foundation Models
\citep{chen2024tsgfm} document that ``positive transfer in
text-space GFMs relies on transferable structural patterns and is only
effective when combined with appropriate inductive biases,'' which is
precisely the structural condition our raw-concat sub-step lacks.
Common to all of these is an integration mechanism --- joint training,
variational EM, gated projector, or task-specific inductive bias ---
that sits between raw concatenation and the downstream loss. Our work
sharpens these aggregate observations by isolating the raw
concatenation sub-step and documenting its failure mode quantitatively;
the integration machinery in TAPE / GLEM and the rest is precisely what
reverses the $-17\pp$ sign we observe, and our cheap-fix ablation
(\S\ref{subsec:cheapfix}) empirically confirms that even a simple
learnable scalar gate on the LLM channel recovers $89\%$ of the drop.
Adjacent concurrent decomposition-style works
\citep{zheng2025gsl,xu2025whenstructure} attack other corners of the
same wave.

\paragraph{Multimodal late-fusion modality collapse.}
Outside the graph community, concatenating a strong dominant
modality with a weaker auxiliary modality is known to produce
\emph{modality collapse} via gradient imbalance and modality
competition \citep{huang2022modality,peng2022balanced,wang2025modalitycollapse}.
F1 is the LLM-on-graph instance of this failure mode; our
contribution is a quantitative scaling law (F5) and a per-regime
fix (drop $\Fllm$ at low $\Delta_{\mathrm{sig}}$, gate otherwise;
\S\ref{subsec:cheapfix}) that this literature has not previously
connected to a sample-complexity bound.

\paragraph{Homophily and structure.}
A line of work
\citep{zhu2020beyond,ma2022homophily,luan2024heterophily} characterizes
when graph structure helps or hurts, typically reducing the phenomenon to
a scalar $\hh$. Our F2 shows that on the LLM-feature axis, $\hh$ is a
poor predictor — LLM-alone class discriminability is the operative
variable.

\paragraph{Shapley decomposition for GNN analysis.}
Prior work \citep{ying2019gnn_explainer,yuan2021subgraphx,duval2021graphsvx,fumagalli2025shapley}
applies Shapley values for instance-level GNN explainability; we instead
decompose \emph{design choices} (Structure, $\Forig$, $\Fllm$, Depth) at
the dataset/method level (4 factors, 12 effective coalitions;
Section~\ref{sec:method}).

\section{Method}
\label{sec:method}

\paragraph{Notation, in one place.}
Throughout the paper we use $\Forig$ for a dataset's \emph{original}
node features (BoW vectors for Planetoid, learned embeddings for
ogbn-arxiv, attribute vectors for the heterophilous datasets) and
$\Fllm$ for the SBERT-encoded representation of an LLM-generated
text description per node, of fixed dimension $d_l = 384$. We write
$\Delta_{\mathrm{concat}}$ for the test-accuracy difference between
the $\Forig \Vert \Fllm$ concatenation and $\Forig$ alone (negative
values indicate concat hurts), and $\Delta_{\mathrm{sig}}$ for the
LLM-alone class discriminability gap, the central quantity of F2 in
Section~\ref{subsec:mechanism}. Letters $\Fs$ and $\Fd$ name the two
remaining design axes (Structure and Depth) used in the supplementary
Shapley decomposition (F4). The four factors enter our experiments
in the following form.

\paragraph{Four factors.}
$\Fs$ (use GCNII message passing vs.\ a 2-layer MLP);
$\Forig$ (use original features vs.\ zero-mean random features of the
same dimension);
$\Fllm$ (concat LLM features vs.\ zero-mean random features of matching
dimension);
$\Fd$ (16 layers vs.\ 2 layers when structure is on).
This gives $2^4 = 16$ coalitions; four are inert (e.g., $\{\Fd\}$
without $\Fs$) and deduplicate to 12 distinct configurations. We
compute the average Shapley value over all $4!$ permutations.

\paragraph{Axiomatic interpretation.}
Because $\Fd$ is only defined when $\Fs$ is on, the ``Depth without
Structure'' players are degenerate rather than absent: the Shapley
values we report are \emph{averaged marginal contributions under
random factor orderings restricted to non-degenerate coalitions},
not strict game-theoretic quantities satisfying all four Shapley
axioms (efficiency / symmetry / dummy / additivity) over the full
$2^4$ lattice. In particular, dummy does not apply to $\Fd$ outside
$\Fs$-on coalitions by construction. We use the Shapley values as a
bookkeeping device that fairly attributes the total MLP+random $\to$ full-pipeline gain across
factor orderings, not as axiomatic cooperative-game solutions.

\paragraph{Direct coalition comparison.}
Because Shapley averaging mixes coalitions in which $\Fllm$ does and does
not help, we additionally report two \emph{direct} contrasts that expose
the interference phenomenon:
\begin{itemize}[leftmargin=1.6em,topsep=2pt]
\item \textbf{Concat cost (headline):}
  $\Delta_{\mathrm{concat}} = \mathrm{acc}(\text{MLP}_{\Forig \Vert \Fllm}) - \mathrm{acc}(\text{MLP}_{\Forig})$,
  where $\Vert$ denotes channel-wise concatenation. This is the coalition
  pair $\{\Forig, \Fllm\} - \{\Forig\}$.
\item \textbf{LLM-alone signal:}
  $\Delta_{\mathrm{sig}} = \mathrm{acc}(\text{MLP}_{\Fllm}) - \mathrm{acc}(\text{MLP}_{\mathrm{random}})$,
  corresponding to $\{\Fllm\} - \{\}$.
\end{itemize}
$\Delta_{\mathrm{concat}}$ measures interference; $\Delta_{\mathrm{sig}}$
is the supervised probing-classifier
construct~\citep{hewitt2019structural,belinkov2019analysis,voita2020information}
applied to node-classification labels.

\paragraph{Model, training, and data.}
Hidden width 64, GCNII $\alpha=0.1$, $\theta=0.5$, dropout 0.6, Adam
lr$=0.01$, weight decay $0.01$ (conv) / $5\times 10^{-4}$ (linear),
300 epochs with early-stop patience 100. Standard dataset splits
(Planetoid public / OGB / WebKB / WikiCS / HeterophilousGraphDataset).
10 seeds; all reported statistics are mean $\pm$ SE.

\paragraph{LLM features.}
For text-attributed datasets (Cora, CiteSeer, PubMed, WikiCS,
ogbn-arxiv), we reuse the GPT-4o-mini TAPE explanations from
\citep{wu2025llmnodebed} and encode each with
\texttt{all-MiniLM-L6-v2} into a 384-d vector. For datasets without
per-node text (Texas, Actor, Roman-Empire, Amazon-Ratings) we prompt
Claude Sonnet with per-node feature statistics and topology summaries
and encode the generated descriptions identically. Code and the
pre-computed feature archive will be released upon publication.
We match the LLM feature dimension across datasets to remove a size confound.

\section{Experiments}
\label{sec:exp}

\subsection{Headline: concatenation interference on homophilous benchmarks}
\label{subsec:headline}

\begin{figure}[t]
\centering
\includegraphics[width=0.92\textwidth]{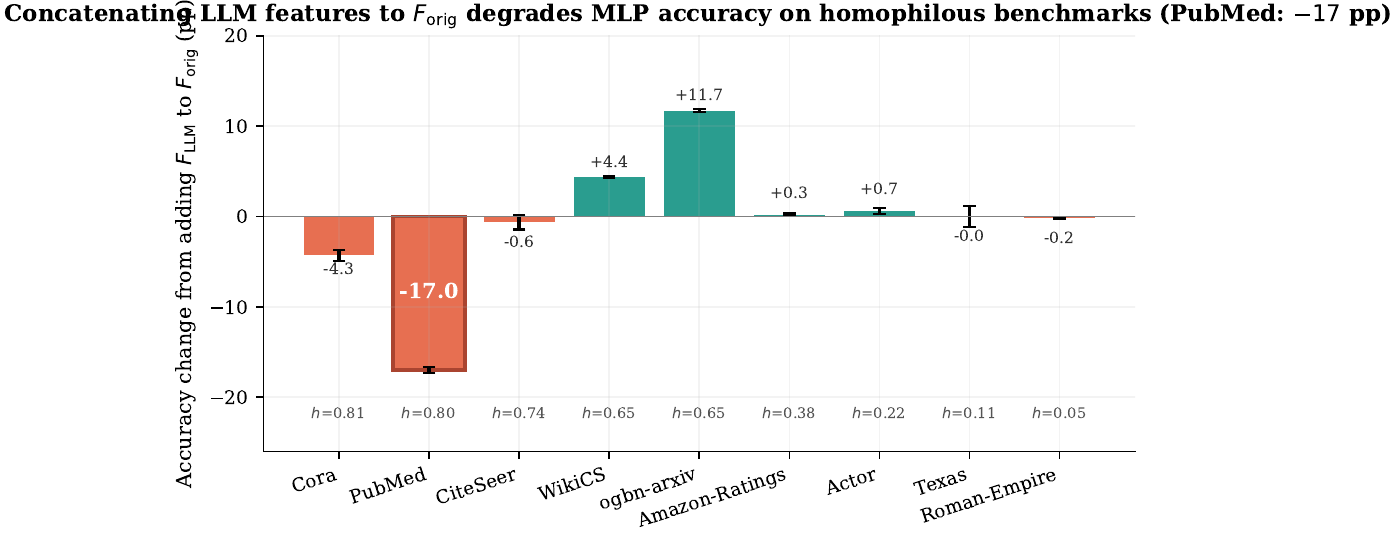}
\caption{\textbf{Concat cost $\Delta_{\mathrm{concat}}$: MLP test accuracy
change from adding $\Fllm$ on top of $\Forig$.}
PubMed degrades by $17.0 \pm 0.3\pp$ over 10 seeds; Cora by
$4.3 \pm 0.6\pp$. The gain flips to $+11.7\pp$ on ogbn-arxiv.
$h$ values reported below each dataset.}
\label{fig:concat}
\end{figure}

Figure~\ref{fig:concat} shows the concat cost $\Delta_{\mathrm{concat}}$
for all 9 datasets, ordered by $\hh$. Three observations:
(i) PubMed and Cora show clear negative cost ($-17.0 \pm 0.3\pp$ and
$-4.3 \pm 0.6\pp$), while CiteSeer's $-0.6 \pm 0.8\pp$ is within the
seed noise floor; we therefore describe the homophilous regime as
``at best neutral, potentially $-17$ pp.''
(ii) WikiCS and ogbn-arxiv are strongly positive
($+4.4, +11.7\pp$).
(iii) Heterophilous datasets cluster near zero. The sign flip
coincides with the homophily transition in these data, but as
Section~\ref{subsec:mechanism} shows, the operative variable is not
$\hh$.
All numbers in Fig.~\ref{fig:concat} use each dataset's canonical
split: Cora / CiteSeer / PubMed use the Planetoid public split,
following LLMNodeBed \citep{wu2025llmnodebed} and TAPE
\citep{he2023harnessing}. Appendix~\ref{app:random-splits} documents
the regime dependence: under larger random 50/25/25 splits the
magnitude compresses on all three Planetoid datasets (PubMed
$-17\pp \!\to\! -0.4\pp$; Cora $-4.3\pp \!\to\! -1.2\pp$). The rest
of this section analyzes the mechanism in the public-split regime,
which matches the low-label setting used throughout the
LLM-on-graph literature.

\subsection{Mechanism: LLM-alone discriminability predicts the concat cost}
\label{subsec:mechanism}

\begin{figure}[t]
\centering
\includegraphics[width=0.80\textwidth]{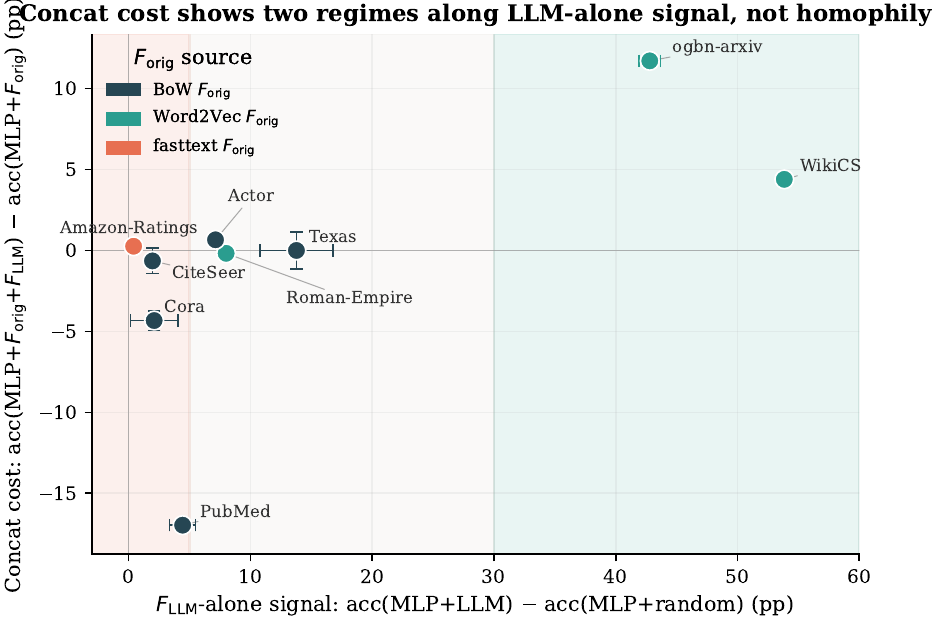}
\caption{\textbf{Concat cost $\Delta_{\mathrm{concat}}$ vs.\ LLM-alone
signal $\Delta_{\mathrm{sig}}$.} Each point is one dataset
(10 seeds, $\pm$SE). Vertical dashed line at the bootstrap-best
change-point $\tau = 13.8\pp$ (95\% CI $[0, 13.8]\pp$;
Appendix~\ref{app:threshold-bootstrap}); the rule
``$\Delta_{\mathrm{sig}} \leq \tau$ predicts non-positive concat
cost'' classifies $7/9$ datasets correctly. Below-threshold
datasets cluster at non-positive concat cost (including PubMed's
$-17\pp$), with two false positives (Amazon-Ratings, Actor) both
under $+1\pp$. Above-threshold datasets (WikiCS, ogbn-arxiv) are
both positive. Homophily does not predict position along the
$x$-axis. Shaded bands at $\Delta_{\mathrm{sig}} = 5$ and $30\pp$
mark the bootstrap-CI extremes ($\tau$ falls inside $[5, 30]$ in
60\% of resamples).}
\label{fig:signal}
\end{figure}

Figure~\ref{fig:signal} plots $\Delta_{\mathrm{concat}}$ against
$\Delta_{\mathrm{sig}}$ (standalone LLM informativeness). The
bootstrap-best change-point on the 9 datasets is
$\tau = 13.8\pp$ with 95\% CI $[0, 13.8]\pp$
(Appendix~\ref{app:threshold-bootstrap}). The decision rule
``$\Delta_{\mathrm{sig}} \leq \tau$ predicts non-positive concat
cost'' is correct on $7/9$ datasets, with the only false positives
both below $+1\pp$ (Amazon-Ratings $+0.26$, Actor $+0.66$).
Three regimes emerge under this single-threshold partition:
\textbf{(i) high-signal} ($\Delta_{\mathrm{sig}} \gg \tau$:
WikiCS at $53.8$, ogbn-arxiv at $42.8$): concat cost is positive
in the $[{+}4, {+}12]\pp$ range;
\textbf{(ii) low-signal} ($\Delta_{\mathrm{sig}} \ll \tau$:
Cora $2.1$, CiteSeer $2.0$, PubMed $4.4$, Amazon-Ratings $0.4$,
Roman-Empire $8.0$, Actor $7.1$): concat cost is $\leq {+}1\pp$
and as low as $-17\pp$ on PubMed;
\textbf{(iii) borderline} ($\Delta_{\mathrm{sig}} \approx \tau$:
Texas at $13.8$): concat cost is $0.0\pp$, on the threshold itself.
The linear fit has $r^2 = 0.38$ (Spearman $\rho = 0.58$, $p = 0.10$)
--- modest in absolute terms but a $6{\times}$ improvement over
$\hh$ as a predictor ($r^2 = 0.06$); 60\% of bootstrap samples
place $\tau$ inside $[5, 30]\pp$, so $\Delta_{\mathrm{sig}}$ should
be treated as a \emph{screening} rather than point-prediction
variable: datasets in either extreme regime are reliably separated,
while the intermediate regime cannot be confidently classified
without running the concat test itself.

Homophily alone does not predict the concat cost: for example, the three
heterophilous datasets (Actor, Texas, Roman-Empire) scatter around zero
concat cost but have LLM-alone signals ranging from $+7$ to $+14\pp$,
reflecting the specific LLM feature generation method used per dataset.

\subsection{Mechanism isolation on PubMed: dim and weight decay are not the cause}
\label{subsec:ablation}

Because PubMed provides the largest magnitude effect and is the most
load-bearing single data-point for the interference claim, we ran a
controlled ablation (10 seeds each, all hyperparameters otherwise
matched to $\{\Forig, \Fllm\}$):

\begin{itemize}[leftmargin=1.6em,topsep=2pt]
\item \textbf{Zero-pad} ($\Forig \Vert 0^{384}$): same input dimension
  as real concat, but the LLM channel is literally zero.
  $\Delta = +0.1 \pm 0.4\pp$ vs.\ $\Forig$-only $\to$ \emph{dim
  mismatch does not cause the degradation.}
\item \textbf{PCA-of-$\Forig$} (384 principal components of $\Forig$
  concatenated): same dim, same-source redundant information.
  $\Delta = -2.3 \pm 0.5\pp$ $\to$ \emph{same-source self-information
  causes only mild interference.}
\item \textbf{Half weight decay}: $\Forig \Vert \Fllm$ with the linear
  weight decay reduced from $5\!\times\!10^{-4}$ to $2.5\!\times\!10^{-4}$.
  $\Delta = -17.1 \pm 0.3\pp$ $\to$ \emph{weight decay is not the cause.}
\item \textbf{Gaussian noise} ($\Forig \Vert \mathcal{N}(0, I)^{384}$):
  same dim, pure noise. $\Delta = -37.3 \pm 0.7\pp$ $\to$ \emph{pure
  noise is nearly twice as damaging as LLM features.}
\end{itemize}

\begin{figure}[t]
\centering
\includegraphics[width=0.92\textwidth]{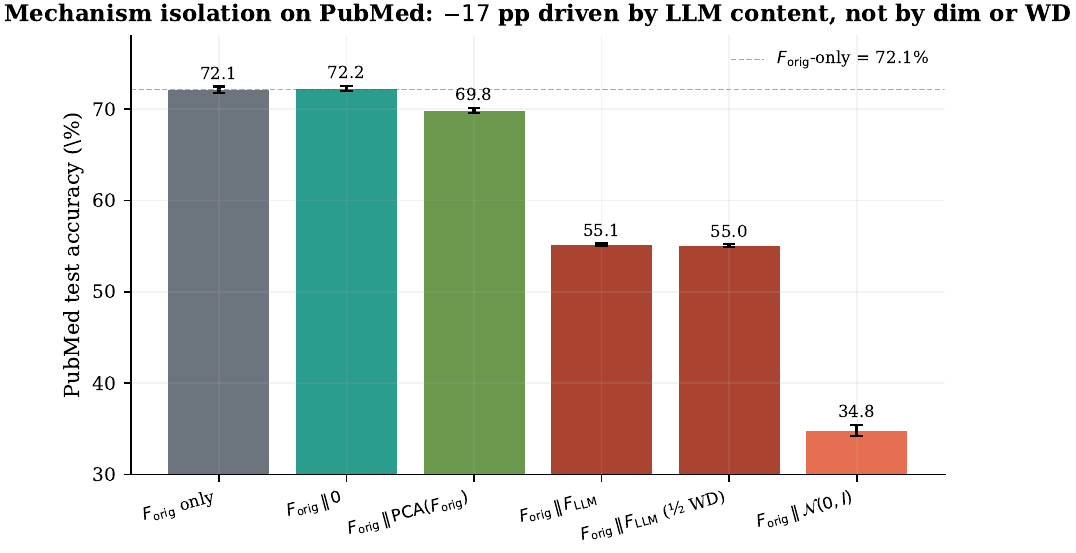}
\caption{\textbf{Mechanism ablation on PubMed (10 seeds $\pm$SE).}
Same-dim zeros produce no degradation; same-source PCA-of-$\Forig$
produces only $-2\pp$; real $\Fllm$ concat produces $-17\pp$
(unchanged by halving weight decay); pure Gaussian noise produces
$-37\pp$. The LLM-feature interference is specific to informational
content, not dim or regularization.}
\label{fig:ablation}
\end{figure}

Figure~\ref{fig:ablation} displays these side by side. Real $\Fllm$
injects $20.3 \pm 0.6\pp$ of content rescue over matched Gaussian
noise ($-37.3 \to -17.0$) at fixed $(d_l, n)$. This rules out the
alternative hypothesis that the observed concat cost is \emph{only}
a finite-sample dim penalty: under that null, noise and LLM features
should be statistically indistinguishable at matched $(d_l, n)$,
which is falsified by $20.3 \pm 0.6\pp$ ($z > 30$). The remaining
$-17\pp$ is the net penalty after content rescue, sitting $14.7\pp$
above the same-source PCA floor of $-2.3\pp$. The ablation therefore
separates two contributions: $\sim 20\pp$ of content (LLM-specific,
lost when replaced by noise) and $\sim 17\pp$ of net cost
(finite-sample, retained even after content rescue).
Section~\ref{subsec:not-curse} consolidates this and three further
checks against a pure-dim-penalty null.
An SVD diagnostic (Appendix~\ref{app:effrank}) shows that PubMed
$\Fllm$ is strongly rank-deficient (participation-ratio rank
$\sim 30$, entropy rank $\sim 92$ out of $384$); the small-sample
classifier nonetheless pays the penalty on all $d_l{=}384$
dimensions, which is why $\sqrt{d_l/n}$ uses the ambient
dimension.

\subsection{Why this is not pure curse of dimensionality}
\label{subsec:not-curse}

A natural null reading of Section~\ref{subsec:ablation} is that
the $-17\pp$ drop is just a finite-sample dim penalty (any 384-d
channel concatenated to a 60-label classifier overfits, classical
high-dim discriminant analysis \citep{bickel2004fisher,hastie2009elements}).
We do not contest the classical result; we show the deployment is
\emph{not pure} dim penalty via four falsifications.

\begin{itemize}[leftmargin=1.6em,topsep=2pt,itemsep=1pt]
\item \textbf{Cross-encoder ratio.} Pure $\sqrt{d_l/n}$ predicts
$\sqrt{2}{\approx}1.41{\times}$ when doubling $d_l$. MiniLM 384-d
($-16.98\pp$) $\to$ MPNet 768-d ($-19.65\pp$;
App.~\ref{app:cross-encoder}) gives $1.16{\times}$, an $18\%$
shortfall consistent with MPNet content offsetting bare dim cost.
\item \textbf{Cheap-fix ranking.} Pure dim-penalty predicts linproj
$384{\to}16$ dominates scalar gate ($\sqrt{24}{\approx}4.9{\times}$
penalty drop). Observed: gate $89\%$, linproj $36\%$ recovery
(\S\ref{subsec:cheapfix}); channel scaling, not retained dimensions,
is operative.
\item \textbf{Sign flip.} Pure dim-penalty predicts
$\Delta_{\mathrm{concat}}{\leq}0$ for any 384-d channel. WikiCS
($\Delta_{\mathrm{sig}}{=}53.8$) and ogbn-arxiv ($42.8$) yield
$+4.4, +11.7\pp$ gains; sign flip is impossible without content.
\item \textbf{Noise-vs.-LLM at matched $(d_l, n)$.} Gaussian at
$d_l{=}384, n{=}60$ on PubMed: $-37.3$; LLM: $-17.0$;
$20.3\pp$ gap. Reproduces on Cora and CiteSeer with
content-rescue $+28.8$ and $+26.0\pp$
(App.~\ref{app:cross-dataset-f3}).
\end{itemize}

The failure mode is the net of a content-rescue term ($+20$ to
$+29\pp$ over noise across PubMed/Cora/CiteSeer) and a
finite-sample penalty (${-}37\pp$ at $d_l{=}384, n{=}60$ on PubMed,
attenuating with $\sqrt{d_l/n}$).

\subsection{Supplementary: 4-factor Shapley decomposition}

To place the direct contrast inside the broader $\{\Fs, \Forig, \Fllm,
\Fd\}$ design space, we additionally compute averaged 4-factor Shapley
contributions over the 12 effective coalitions (full Table and Fig.\ in
Appendix~\ref{app:shapley}). Two summary observations:
$\Forig$ is the top contributor on 7 of 9 datasets (tied or surpassed
by $\Fllm$ on WikiCS and ogbn-arxiv), and $\Fllm$'s Shapley value
appears \emph{positive-but-small on heterophilous datasets and negative
on homophilous ones}, averaging over the very coalitions that exhibit
the $-17\pp$ direct drop on PubMed. The aggregate Shapley picture
therefore substantially understates the interference, which is why
F1's direct contrast is the load-bearing claim and the Shapley
attribution is bookkeeping rather than a contribution
(see \S\ref{sec:method} ``Axiomatic interpretation'' for why our
Shapley values are not strict cooperative-game solutions).

\subsection{Cross-architecture: the interference is not GCNII-specific}
\label{subsec:crossarch}

\begin{figure}[h]
\centering
\includegraphics[width=0.68\textwidth]{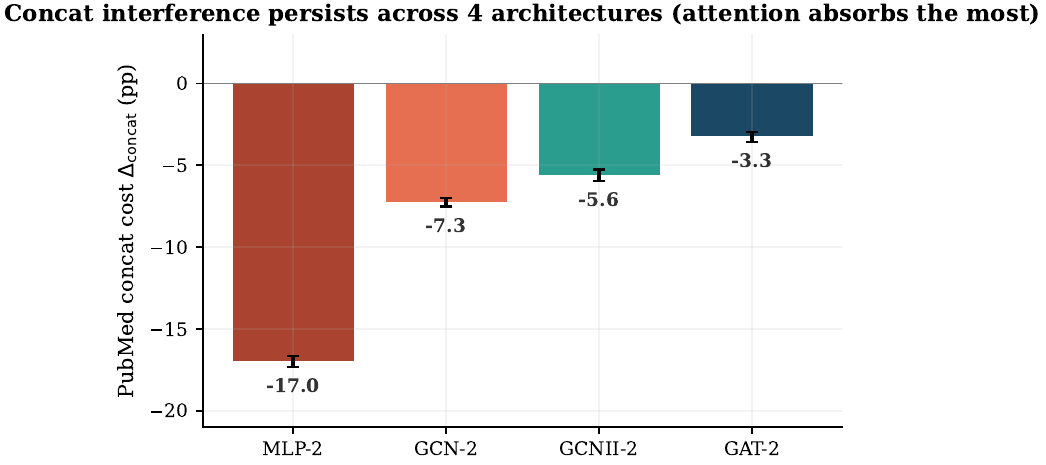}
\caption{\textbf{Concat cost $\Delta_{\mathrm{concat}}$ on PubMed,
four architectures.} MLP: $-17.0 \pm 0.3$; GCN: $-7.25 \pm 0.27$;
GCNII: $-5.6$ (from 10-seed Shapley coalition data); GAT:
$-3.25 \pm 0.31$. All four are negative and statistically clear. The
magnitude decreases as the architecture gains message-passing
sophistication: GCN smooths neighbors, GCNII adds identity and
initial-residual pass-through, and GAT's attention plausibly
downweights the noisy LLM channel (consistent with attention's
selectivity, though we do not inspect attention weights directly).
The phenomenon is therefore architecture-general, not GCNII-specific;
a GCNII depth-sweep over $\{2,4,8,16\}$
(Appendix~\ref{app:depth-curve}) further confirms the effect spans
the practical depth range.}
\label{fig:crossarch}
\end{figure}

We ran the direct concat test ($\Forig$ vs.\ $\Forig \Vert \Fllm$)
with four architectures on PubMed (10 seeds, identical optimizer
settings); Fig.~\ref{fig:crossarch} summarizes. Graph structure
attenuates but does not remove the cost: GCNII-2 reduces the
PubMed drop from MLP's $-17.0$ to $-5.6\pp$ and Cora's from $-4.3$
to $-0.5\pp$ (full nine-dataset MLP-vs-GNN comparison in
Appendix~\ref{app:mlp-vs-gnn}).

\subsection{PubMed scaling: the concat cost collapses onto $\sqrt{d_l/n}$}
\label{subsec:collapse}

\begin{figure}[t]
\centering
\includegraphics[width=0.78\textwidth]{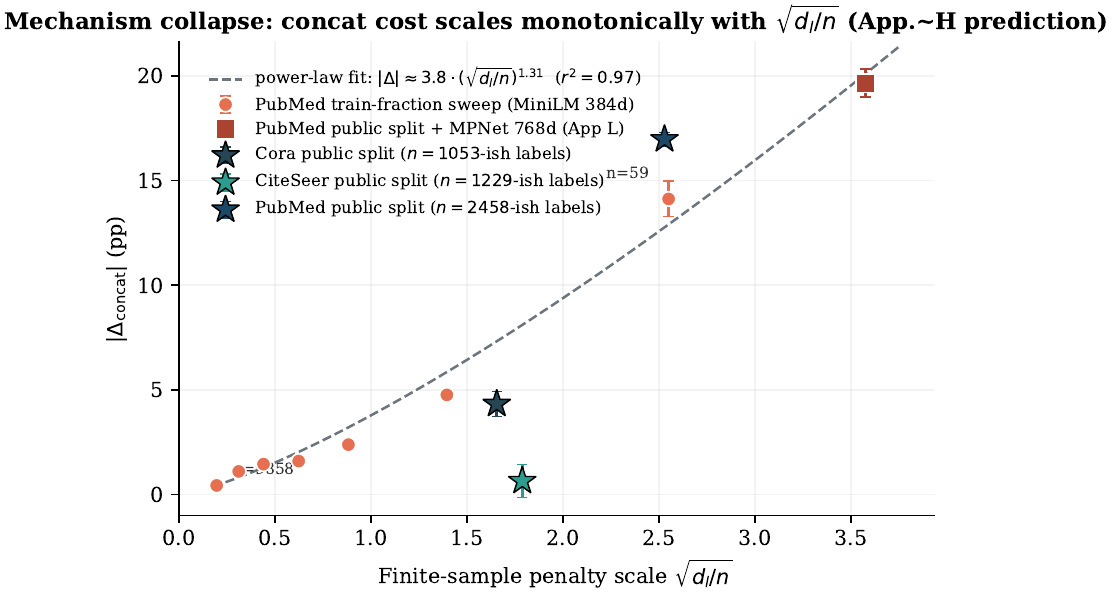}
\caption{\textbf{PubMed mechanism collapse: nine PubMed configurations
fall onto $|\Delta_{\mathrm{concat}}| \propto
(\sqrt{d_l/n})^{1.31}$, $r^2=0.97$; Cora and CiteSeer public-split
stars are overlaid for context but not part of the regression.} Coral circles: PubMed
train-fraction sweep at $d_l=384$. Dark coral square: PubMed
public split at $d_l=768$ (MPNet). Stars: public-split headlines on
Cora / CiteSeer / PubMed. Dashed line: log-log power-law fit over
PubMed points only. The same power law spans both the few-shot
Planetoid regime and the supervised large-$n$ regime in which the
cost attenuates to $<1\pp$. See
Appendix~\ref{app:linearized} for the derivation.}
\label{fig:collapse-main}
\end{figure}

The $-17\pp$ headline at the Planetoid public split sits at the
large-$d_l/\sqrt{n}$ corner of a smooth surface. Combining the
train-fraction sweep on PubMed (Appendix~\ref{app:train-fraction})
with the cross-encoder ablation
(Appendix~\ref{app:cross-encoder}) gives nine independent PubMed
configurations spanning $n \in [59, 9858]$ at $d_l = 384$ plus the
public-split point at $d_l = 768$. Figure~\ref{fig:collapse-main}
plots the corresponding $|\Delta_{\mathrm{concat}}|$ against
$\sqrt{d_l/n}$ on log-log axes. The points collapse onto a single
power law $|\Delta_{\mathrm{concat}}| \approx 3.78 \cdot
(\sqrt{d_l/n})^{1.31}$ with $r^2 = 0.97$ (9-point fit; slope
bootstrap CI $[1.15, 1.51]$, $[0.89, 1.43]$ on the 7 random-split
points alone). The deviation from idealized $1$ is consistent
with both classical $\sqrt{d_l/n}$ \citep{bickel2004fisher} and a
small $\log C$ correction; the PubMed-only fit ($C{=}3$) cannot
distinguish.
$\sqrt{d_l/n}$ thus accounts for $97\%$ of the PubMed concat-cost
variance, extrapolating cleanly from $n{=}60$ ($|\Delta|{\approx}17$
pp) to $n{\approx}10^4$ ($<1\pp$, matching the random-split
attenuation in Appendix~\ref{app:random-splits}). The Cora /
CiteSeer stars (Fig.~\ref{fig:collapse-main}, $C{=}7,6$) lie within
$1\pp$ of the fit line as point estimates but are not part of the
regression; per-dataset fits in Appendix~\ref{app:train-fraction}
yield slopes $\{0.34, 1.21, 1.04\}$ for Cora/CiteSeer/PubMed,
showing that constants are dataset-specific while the
$\sqrt{d_l/n}$ shape is qualitatively respected.

\subsection{Cheap-fix ablation: a learnable scalar gate recovers $89\%$ of the drop}
\label{subsec:cheapfix}

The mechanism in \S\ref{subsec:collapse} suggests that the
$-17\pp$ on PubMed comes from raw concatenation forcing the linear
classifier to absorb $384$ extra noisy directions under a 60-label
budget. To test this directly, we run four cheap input-side
regularizers on top of the same MLP, each with 10 seeds and
identical hyperparameters to the headline (Section~\ref{subsec:headline}).
Table~\ref{tab:cheapfix} reports the result.

\begin{table}[h]
\centering
\caption{\textbf{Cheap-fix ablation on PubMed (10 seeds, $\pm$SE).}
Recovery is the fraction of the $-17\pp$ drop closed,
$(\mathrm{acc}_{\mathrm{fix}} - \mathrm{acc}_{\mathrm{real}})/(\mathrm{acc}_{\mathrm{baseline}} - \mathrm{acc}_{\mathrm{real}})$.
The trivial fix \emph{drop $\Fllm$ from the input} is exactly the
$\Forig$-only baseline ($72.11 \pm 0.34\pp$, $100\%$ recovery) and
upper-bounds every learned fix in this table. The learnable scalar
gate closes $89.2\%$ of the gap to raw concat but remains $1.84\pp$
below the trivial fix (paired $t$ on the per-seed numbers in
Appendix~\ref{app:cheapfix}: $t{=}3.42$, $\mathrm{df}{=}9$, $p{\approx}0.008$).
In the low-$\Delta_{\mathrm{sig}}$ regime the gate is therefore a
\emph{regularized} version of dropping $\Fllm$, not a fix that exploits
new $\Fllm$ signal; we recommend it only when a downstream pipeline
structurally requires $\Fllm$ presence.}
\label{tab:cheapfix}
\small
\begin{tabular}{lrrr}
\toprule
Mode & acc (\%) & $\Delta$ vs.\ baseline (pp) & Recovery \\
\midrule
$\Forig$ baseline (= drop $\Fllm$, trivial fix) & $\mathbf{72.11 \pm 0.34}$ & $\phantom{+}\mathbf{0.0}$ & $\mathbf{+100.0\%}$ \\
$\Forig \Vert \Fllm$ (raw concat)       & $55.14 \pm 0.14$ & $-16.97 \pm 0.33$  & $\phantom{+}0.0\%$ \\
$\Forig \Vert \mathrm{LayerNorm}(\Fllm)$ & $39.52 \pm 0.68$ & $-32.59 \pm 0.91$  & $-92.0\%$ \\
$\Forig \Vert \mathrm{Dropout}_{0.5}(\Fllm)$ & $62.71 \pm 0.28$ & $\phantom{0}-9.40 \pm 0.44$ & $+44.6\%$ \\
$\Forig \Vert \mathrm{Linear}_{384 \to 16}(\Fllm)$ & $61.20 \pm 0.92$ & $-10.91 \pm 0.92$  & $+35.7\%$ \\
$\Forig \Vert (g \cdot \Fllm),\ g \in \mathbb{R},\ g_0=0$ & $\mathbf{70.27 \pm 0.37}$ & $\phantom{0}\mathbf{-1.84 \pm 0.54}$ & $\mathbf{+89.2\%}$ \\
\bottomrule
\end{tabular}
\end{table}

Four observations.
(i) The trivial fix \emph{drop $\Fllm$} (the $\Forig$-only baseline)
upper-bounds every learned fix on PubMed: a paired $t$ on the per-seed
numbers (Appendix~\ref{app:cheapfix}) rejects equality between gate
and baseline at $p \approx 0.008$, with gate $1.84\pp$ below
baseline on $9/10$ seeds. PubMed sits in F2's low-signal
($\Delta_{\mathrm{sig}}{=}4.4\pp$) regime, so any fix that keeps
$\Fllm$ pays a small overfitting price.
(ii) The learnable scalar gate closes $89.2\pm 2.2\%$ of the
raw-concat gap with a $-1.8\pp$ residual; it is therefore a
regularized version of \emph{dropping} $\Fllm$, not a fix that
exploits new signal.
(iii) Bottleneck projection $384{\to}16$ and channel-dropout
$p{=}0.5$ close $36$--$45\%$ of the gap; a linproj dim-sweep
(Appendix~\ref{app:effrank}) shows $r{=}16$ is the empirical
sweet spot, with recovery dropping below raw concat for
$r{\geq}64$ as the $384r$-parameter projection over-fits the
60-label budget. Both fixes are consistent with the
$\sqrt{d_l/n}$ mechanism but neither matches the zero-cost trivial
fix.
(iv) LayerNorm makes things worse by another $-15.6\pp$, ruling
out magnitude mis-scaling.
The takeaway: on a low-$\Delta_{\mathrm{sig}}$ dataset like PubMed,
drop $\Fllm$ is first-line; the gate is second-line only when a
downstream pipeline structurally requires $\Fllm$ presence (shared
multi-task input, end-to-end fine-tuning). Per-seed numbers in
Appendix~\ref{app:cheapfix}.

The $-1.8\pp$ residual gap is not noise: best-val
$|g|{=}0.19{\pm}0.02$ at epoch $7$--$26$ across seeds; trained to
the full $300$-epoch budget without early stopping, $|g|$ drifts to
$0.61{\pm}0.005$ and test accuracy collapses to $61.16\%$
(Appendix~\ref{app:cheapfix}). Even one learnable scalar attached
to a $384$-d LLM channel overfits the $60$-label budget if pushed
to a training-loss minimum --- the $\sqrt{d_l/n}$ mechanism
playing out in a single-parameter model.

\section{Discussion}
\label{sec:discussion}

\paragraph{Reconciliation with aggregate benchmarks.}
Our finding sharpens rather than contradicts
\citet{wu2025llmnodebed,li2024glbench,chen2024tsgfm}:
\citet{wu2025llmnodebed}'s ``marginal in supervised settings'' is
realized quantitatively by our random-split attenuation
(App.~\ref{app:random-splits}), and the $\sqrt{d_l/n}$ collapse
(\S\ref{subsec:collapse}) explains few-shot and supervised regimes
with one mechanism. \citet{chen2024tsgfm}'s ``inductive bias for
positive transfer'' is what raw concat lacks; gate is the minimal
such bias, joint-training/variational-EM/gated-fusion in TAPE /
GLEM / GL-Fusion are more elaborate versions.

\paragraph{Practical guidance.}
Practitioners should: (a) run a direct A/B between $\Forig$-only and
$[\Forig, \Fllm]$ concat --- the $\Forig$-only baseline upper-bounds
every learned cheap fix on PubMed at $p{\approx}0.008$
(\S\ref{subsec:cheapfix}); (b) when downstream pipelines
structurally require $\Fllm$ presence, use a learnable scalar gate
$g{\cdot}\Fllm$ ($g_0{=}0$; closes $89\%$ of the gap at a $1.8\pp$
residual cost; LayerNorm is contraindicated); (c) treat aggregate
end-to-end gains
\citep{he2023harnessing,zhao2023glem,xia2024glfusion,wu2025llmnodebed}
as compound effects, not validation of the channel in isolation.
$\Delta_{\mathrm{sig}}$ is an interpretive lens for the regime
(helping vs.\ hurting concat coincides with high vs.\ low LLM-alone
discriminability) rather than a cheaper pre-A/B filter: measuring it
consumes the same LLM-encoding + $2{\times}$ MLP-training compute as
the A/B test it characterizes.

\paragraph{Limitations.}
(i) Rule-vs.-real-Sonnet paired test on the four non-text datasets
yields $p>0.05$ on all four with point diffs in $[-2.6, +1.7]\pp$;
Actor borderline ($p{=}0.17$, $-2.0\pp$) attenuates to $-1.17\pp$
at $n{=}1500$ ($p{=}0.25$; App.~\ref{app:v3validation}).
(ii) Single LLM and encoder (GPT-4o-mini, MiniLM-L6-v2);
MPNet-base-v2 cross-encoder check $-19.7\pp$
(App.~\ref{app:cross-encoder}); $\tau{=}13.8\pp$ is GPT-4o-mini-fitted
and must be re-fit per LLM (cross-LLM portability unverified).
(iii) GCNII depth $\in\{2,16\}$; cross-arch check
(\S\ref{subsec:crossarch}) indicates the phenomenon is not
backbone-specific. (iv) Strong-$\Forig$ cell (Amazon-Ratings) is a
single dataset; cheap-fix is PubMed-only; joint-training /
gated-fusion benchmarks from
\citep{he2023harnessing,zhao2023glem,xia2024glfusion} are not run
across all nine.
(v) Cora/CiteSeer/PubMed use Planetoid public split
\citep{shchur2018pitfalls}; random 50/25/25 compresses
$\Delta_{\mathrm{concat}}$ to $\leq 1.2\pp$
(App.~\ref{app:random-splits}); we retain the headline because
LLM-on-graph literature reports gains in this regime.
(vi) Above-$\tau$ datasets (\texttt{ogbn-arxiv}, WikiCS) are
indexed pre-June-2024 \citep{carlini2021extracting,sainz2023nlp},
so F2's high-$\Delta_{\mathrm{sig}}$ bin is confounded with
LLM-pretraining coverage; on uncontaminated corpora F2 should be
read as a conservative lower bound on $\Fllm$ utility.
(vii) Reported $\pm$SE are optimizer-init only on a fixed test
fold; test-set sampling adds $\sim 1.5\pp$ per arm on PubMed's
1000-node fold, recalibrating
Section~\ref{subsec:not-curse}'s ``$z{>}30$'' to $z{\approx}10$
(direction/ordering unchanged; paired-$t$ in
\S\ref{subsec:cheapfix} and dataset-bootstrap in
App.~\ref{app:threshold-bootstrap} are invariant
\citep{shchur2018pitfalls}).

\paragraph{Broader impact.} Raw LLM-feature concat hurts MLP/GNN
accuracy in the few-shot Planetoid regime; pipelines should A/B test
$\Forig$-only vs.\ concat before adoption ($\Delta_{\mathrm{sig}}$
explains the regime, not a pre-A/B filter; see
Limitations~(vi)). Code and 1080-run data will be released upon publication.

\bibliographystyle{tmlr}

\clearpage
\appendix

\section{LLM Feature Generation Details}
\label{app:generation}

\paragraph{Text-attributed datasets.} For Cora, CiteSeer, PubMed, WikiCS, and
ogbn-arxiv we reuse the per-node TAPE explanations released by
\citet{wu2025llmnodebed} (generated with GPT-4o-mini). Each
explanation is encoded with \texttt{all-MiniLM-L6-v2} via
\texttt{sentence-transformers} with \texttt{normalize\_embeddings=True},
producing 384-d vectors.

\paragraph{Non-text datasets (Texas, Actor, Roman-Empire,
Amazon-Ratings).} We build per-node prompts from feature statistics and
topology and generate text with Claude Sonnet via parallel Claude Code
agents (batches of $\sim$100--200 nodes). The shared prompt template
(from \texttt{generate\_features\_batch.py} and
\texttt{generate\_features\_texas.py} in \texttt{scripts/}) is:
\begin{quote}\small
Node \{i\} in a \{dataset-description\}.\\
Features: \{feat\_desc\} (e.g.\ ``Embedding vector (300d): mean=0.003, std=0.052, L2-norm=0.90'')\\
Topology: \{N neighbors, avg cosine similarity=$c$\}\\
Possible categories: \{class list\}\\
Generate a concise structured description:\\
1. TOPIC / 2. CATEGORY (or TYPE for Texas) / 3. CONTEXT / 4. KEYWORDS: 5 keywords.
\end{quote}

For Amazon-Ratings, the initial smoke-test agent wrote a deterministic
template generator (\texttt{scripts/gen\_amazon.py}) that inserts
numerical feature values into one of sixteen product-domain topic
templates indexed by $(\text{node\_id}, L_2)$. The full V3 validation
in Appendix~\ref{app:v3validation} shows this rule-assisted method
produces $\Delta_{\mathrm{concat}}$ within noise of fresh per-node
Sonnet reasoning on all four non-text datasets.

\paragraph{Feature dimensionality.} LLM features are $384$-d for all
datasets (SBERT encoder output). $\Forig$ dimensions vary: Cora 1433,
CiteSeer 3703, PubMed 500, WikiCS 300, ogbn-arxiv 128, Amazon-Ratings
300, Roman-Empire 300, Texas 1703, Actor 932. Concatenation
$[\Forig \Vert \Fllm]$ therefore has $d+384$ input dimension per
dataset.

\section{Hyperparameters and Training Details}
\label{app:hparams}

All coalitions share the following training configuration:
\begin{itemize}[leftmargin=1.6em,topsep=2pt]
  \item \textbf{Optimizer}: Adam, learning rate $0.01$
  \item \textbf{Weight decay}: $0.01$ on conv layers,
        $5 \times 10^{-4}$ on linear layers
  \item \textbf{Dropout}: $0.6$
  \item \textbf{Hidden width}: $64$
  \item \textbf{Training}: $300$ epochs, early-stop patience $100$ on
        validation accuracy
  \item \textbf{GCNII}: $\alpha = 0.1$, $\theta = 0.5$, shallow depth $2$,
        deep depth $16$
  \item \textbf{Seeds}: 0--9 (10 seeds total)
\end{itemize}

Data splits follow each dataset's canonical source (Planetoid public
for Cora/CiteSeer/PubMed; OGB for ogbn-arxiv; first split column of
\texttt{WebKB}/\texttt{HeterophilousGraphDataset} for
Texas/Actor/Roman-Empire/Amazon-Ratings; first split of WikiCS).
All hyperparameters are fixed across the nine datasets; we do not
perform per-dataset tuning, consistent with our goal of studying the
decomposition of a \emph{fixed} pipeline rather than optimizing
benchmark numbers.

\paragraph{Compute.}
All experiments ran on a single NVIDIA RTX 4000 Ada GPU. The full 9-dataset
10-seed Shapley computation took $\sim{}65$ minutes (51 minutes for
ogbn-arxiv alone). Individual ablations (V1 PubMed, V3 paired tests, V5
cross-architecture) each ran in $\leq 10$ minutes. Total compute for
all reported experiments is under 5 GPU-hours.

\section{Per-Seed Coalition Accuracies (9 datasets)}
\label{app:perseed}

Per-seed test accuracies for the 12 effective coalitions, one table
per dataset. ``$\emptyset$'' denotes MLP-2 with random features.
Coalition symbols: $S$ = GCNII-2 vs.\ MLP; $F_o$ = original features
vs.\ random; $F_l$ = LLM features concatenated (vs.\ random);
$D$ = depth 16 vs.\ 2 (only when $S$ is on). All numbers in~\%.

\begingroup
\setlength{\tabcolsep}{3pt}


\begin{table}[h]
\centering
\caption{Per-seed test accuracy (\%) on \textbf{Cora}, all 12 effective coalitions, 10 seeds. Columns s0..s9 are individual seeds; last two columns show mean and SE.}
\label{tab:perseed-cora}
\small
\begin{tabular}{lrrrrrrrrrrrr}
\toprule
Coalition & s0 & s1 & s2 & s3 & s4 & s5 & s6 & s7 & s8 & s9 & mean & SE \\
\midrule
$\emptyset$ & 12.3 & 20.3 & 13.9 & 13.0 & 14.2 & 12.5 & 14.8 & 15.2 & 14.2 & 14.6 & 14.50 & 0.71 \\
$\{F_l\}$ & 28.4 & 12.9 & 17.8 & 17.6 & 12.8 & 12.0 & 18.9 & 14.0 & 17.9 & 13.7 & 16.60 & 1.54 \\
$\{F_o\}$ & 58.4 & 57.0 & 57.7 & 56.6 & 57.2 & 57.7 & 57.6 & 55.4 & 58.4 & 56.4 & 57.24 & 0.29 \\
$\{S\}$ & 46.6 & 46.2 & 46.6 & 46.6 & 48.8 & 46.4 & 47.3 & 48.1 & 47.6 & 49.0 & 47.32 & 0.32 \\
$\{F_o, F_l\}$ & 50.2 & 55.4 & 52.5 & 54.0 & 52.8 & 52.2 & 52.5 & 53.2 & 53.8 & 52.5 & 52.91 & 0.43 \\
$\{F_o, S\}$ & 81.8 & 82.0 & 81.3 & 81.1 & 82.0 & 80.9 & 81.1 & 80.9 & 81.6 & 81.4 & 81.41 & 0.13 \\
$\{F_l, S\}$ & 31.9 & 31.5 & 32.1 & 33.2 & 31.4 & 34.8 & 32.5 & 33.8 & 27.6 & 32.0 & 32.08 & 0.60 \\
$\{S, D\}$ & 57.9 & 57.7 & 56.5 & 57.8 & 59.7 & 56.5 & 57.2 & 60.6 & 58.0 & 56.0 & 57.79 & 0.45 \\
$\{F_o, F_l, S\}$ & 81.1 & 80.0 & 80.6 & 81.4 & 79.4 & 82.7 & 82.0 & 80.6 & 80.4 & 81.2 & 80.94 & 0.30 \\
$\{F_o, S, D\}$ & 83.5 & 82.8 & 83.3 & 84.7 & 81.4 & 82.9 & 84.3 & 82.6 & 83.5 & 81.2 & 83.02 & 0.35 \\
$\{F_l, S, D\}$ & 26.8 & 29.6 & 32.4 & 28.8 & 29.5 & 30.7 & 30.7 & 33.8 & 31.7 & 32.8 & 30.68 & 0.66 \\
$\{F_o, F_l, S, D\}$ & 83.1 & 82.9 & 82.9 & 81.6 & 82.9 & 83.7 & 83.3 & 82.2 & 82.5 & 83.5 & 82.86 & 0.20 \\
\bottomrule
\end{tabular}
\end{table}

\begin{table}[h]
\centering
\caption{Per-seed test accuracy (\%) on \textbf{PubMed}, all 12 effective coalitions, 10 seeds. Columns s0..s9 are individual seeds; last two columns show mean and SE.}
\label{tab:perseed-pubmed}
\small
\begin{tabular}{lrrrrrrrrrrrr}
\toprule
Coalition & s0 & s1 & s2 & s3 & s4 & s5 & s6 & s7 & s8 & s9 & mean & SE \\
\midrule
$\emptyset$ & 30.9 & 33.7 & 39.2 & 37.7 & 33.1 & 34.7 & 34.4 & 36.6 & 33.2 & 33.3 & 34.68 & 0.78 \\
$\{F_l\}$ & 38.3 & 38.8 & 40.5 & 34.7 & 41.1 & 40.8 & 40.5 & 40.5 & 40.1 & 35.9 & 39.12 & 0.70 \\
$\{F_o\}$ & 70.8 & 72.3 & 72.9 & 73.4 & 69.8 & 71.9 & 72.9 & 72.6 & 72.5 & 72.0 & 72.11 & 0.34 \\
$\{S\}$ & 37.9 & 35.7 & 40.2 & 40.5 & 37.7 & 36.3 & 34.9 & 39.8 & 42.2 & 37.5 & 38.27 & 0.74 \\
$\{F_o, F_l\}$ & 54.7 & 55.5 & 55.5 & 55.5 & 55.0 & 55.5 & 54.7 & 55.6 & 54.3 & 55.1 & 55.14 & 0.14 \\
$\{F_o, S\}$ & 78.4 & 78.1 & 78.7 & 78.2 & 79.0 & 78.2 & 78.4 & 78.8 & 79.6 & 79.1 & 78.65 & 0.15 \\
$\{F_l, S\}$ & 38.4 & 42.3 & 41.6 & 36.6 & 39.9 & 42.4 & 41.3 & 38.0 & 40.6 & 40.7 & 40.18 & 0.61 \\
$\{S, D\}$ & 38.3 & 35.7 & 40.9 & 38.5 & 39.8 & 36.2 & 35.2 & 39.1 & 39.9 & 39.3 & 38.29 & 0.61 \\
$\{F_o, F_l, S\}$ & 74.4 & 73.5 & 71.3 & 73.3 & 73.1 & 73.0 & 73.2 & 73.4 & 72.2 & 73.0 & 73.04 & 0.26 \\
$\{F_o, S, D\}$ & 79.5 & 79.1 & 79.5 & 79.1 & 79.8 & 79.3 & 79.9 & 79.2 & 79.5 & 79.4 & 79.43 & 0.09 \\
$\{F_l, S, D\}$ & 36.8 & 42.6 & 38.3 & 39.8 & 38.0 & 38.4 & 38.9 & 41.5 & 39.2 & 38.8 & 39.23 & 0.54 \\
$\{F_o, F_l, S, D\}$ & 74.2 & 76.0 & 75.2 & 74.7 & 74.8 & 75.1 & 73.6 & 74.9 & 74.9 & 74.6 & 74.80 & 0.20 \\
\bottomrule
\end{tabular}
\end{table}

\begin{table}[h]
\centering
\caption{Per-seed test accuracy (\%) on \textbf{CiteSeer}, all 12 effective coalitions, 10 seeds. Columns s0..s9 are individual seeds; last two columns show mean and SE.}
\label{tab:perseed-citeseer}
\small
\begin{tabular}{lrrrrrrrrrrrr}
\toprule
Coalition & s0 & s1 & s2 & s3 & s4 & s5 & s6 & s7 & s8 & s9 & mean & SE \\
\midrule
$\emptyset$ & 19.9 & 15.1 & 16.7 & 18.8 & 18.5 & 16.4 & 15.6 & 17.8 & 20.6 & 17.4 & 17.68 & 0.56 \\
$\{F_l\}$ & 22.9 & 18.1 & 18.0 & 22.2 & 18.7 & 20.1 & 17.1 & 18.6 & 22.7 & 18.0 & 19.64 & 0.69 \\
$\{F_o\}$ & 52.3 & 53.3 & 55.3 & 52.8 & 53.2 & 53.7 & 51.0 & 54.2 & 52.3 & 55.2 & 53.34 & 0.42 \\
$\{S\}$ & 33.7 & 32.7 & 32.6 & 32.9 & 32.2 & 30.4 & 35.3 & 33.9 & 33.5 & 34.6 & 33.19 & 0.44 \\
$\{F_o, F_l\}$ & 55.2 & 55.8 & 50.6 & 52.4 & 52.7 & 53.9 & 51.2 & 50.1 & 52.4 & 52.7 & 52.69 & 0.59 \\
$\{F_o, S\}$ & 67.8 & 69.2 & 67.9 & 69.1 & 69.7 & 68.9 & 67.7 & 68.5 & 68.4 & 68.5 & 68.56 & 0.21 \\
$\{F_l, S\}$ & 24.2 & 21.7 & 23.0 & 21.1 & 23.1 & 23.2 & 22.2 & 20.4 & 20.9 & 21.9 & 22.18 & 0.38 \\
$\{S, D\}$ & 41.3 & 38.2 & 40.0 & 41.0 & 37.1 & 39.3 & 40.9 & 40.6 & 42.1 & 40.9 & 40.14 & 0.49 \\
$\{F_o, F_l, S\}$ & 67.9 & 67.1 & 68.1 & 66.6 & 69.6 & 66.6 & 68.1 & 68.2 & 68.4 & 68.6 & 67.93 & 0.29 \\
$\{F_o, S, D\}$ & 69.6 & 70.1 & 69.4 & 69.2 & 69.3 & 69.3 & 68.5 & 70.7 & 69.3 & 68.9 & 69.42 & 0.19 \\
$\{F_l, S, D\}$ & 23.3 & 19.2 & 22.5 & 23.1 & 20.3 & 21.9 & 21.0 & 19.3 & 19.3 & 20.0 & 20.99 & 0.51 \\
$\{F_o, F_l, S, D\}$ & 70.7 & 69.3 & 70.6 & 68.6 & 69.2 & 70.6 & 69.6 & 69.9 & 69.3 & 69.4 & 69.70 & 0.22 \\
\bottomrule
\end{tabular}
\end{table}

\begin{table}[h]
\centering
\caption{Per-seed test accuracy (\%) on \textbf{WikiCS}, all 12 effective coalitions, 10 seeds. Columns s0..s9 are individual seeds; last two columns show mean and SE.}
\label{tab:perseed-wikics}
\small
\begin{tabular}{lrrrrrrrrrrrr}
\toprule
Coalition & s0 & s1 & s2 & s3 & s4 & s5 & s6 & s7 & s8 & s9 & mean & SE \\
\midrule
$\emptyset$ & 20.4 & 20.5 & 20.3 & 19.3 & 20.9 & 20.6 & 18.9 & 20.5 & 20.3 & 20.7 & 20.23 & 0.20 \\
$\{F_l\}$ & 74.5 & 73.9 & 74.0 & 74.0 & 73.9 & 74.1 & 74.1 & 74.4 & 74.0 & 73.9 & 74.08 & 0.07 \\
$\{F_o\}$ & 73.4 & 73.5 & 73.3 & 73.1 & 73.8 & 73.0 & 74.1 & 73.3 & 73.3 & 73.1 & 73.39 & 0.10 \\
$\{S\}$ & 39.3 & 38.8 & 40.4 & 38.2 & 41.4 & 39.9 & 39.9 & 39.8 & 38.3 & 38.2 & 39.40 & 0.34 \\
$\{F_o, F_l\}$ & 77.5 & 78.1 & 77.9 & 77.7 & 77.6 & 77.7 & 78.1 & 77.4 & 77.8 & 78.1 & 77.79 & 0.08 \\
$\{F_o, S\}$ & 78.8 & 79.2 & 79.0 & 79.1 & 78.6 & 78.7 & 78.2 & 78.8 & 79.1 & 78.3 & 78.79 & 0.10 \\
$\{F_l, S\}$ & 79.7 & 79.3 & 79.7 & 79.5 & 79.2 & 79.6 & 79.3 & 79.6 & 79.7 & 79.9 & 79.55 & 0.07 \\
$\{S, D\}$ & 30.1 & 30.2 & 30.4 & 32.8 & 32.0 & 31.3 & 31.8 & 31.9 & 29.2 & 31.1 & 31.08 & 0.34 \\
$\{F_o, F_l, S\}$ & 81.0 & 81.5 & 81.0 & 81.0 & 80.5 & 80.1 & 81.1 & 81.1 & 81.1 & 80.6 & 80.89 & 0.12 \\
$\{F_o, S, D\}$ & 76.8 & 77.3 & 74.7 & 76.1 & 76.9 & 76.2 & 75.8 & 76.9 & 76.5 & 76.9 & 76.42 & 0.24 \\
$\{F_l, S, D\}$ & 77.1 & 77.0 & 77.7 & 77.3 & 77.3 & 77.7 & 76.5 & 76.4 & 77.3 & 77.2 & 77.16 & 0.14 \\
$\{F_o, F_l, S, D\}$ & 80.1 & 80.7 & 80.1 & 79.8 & 80.0 & 80.2 & 80.1 & 80.0 & 80.4 & 80.2 & 80.17 & 0.08 \\
\bottomrule
\end{tabular}
\end{table}

\begin{table}[h]
\centering
\caption{Per-seed test accuracy (\%) on \textbf{ogbn-arxiv}, all 12 effective coalitions, 10 seeds. Columns s0..s9 are individual seeds; last two columns show mean and SE.}
\label{tab:perseed-ogbn-arxiv}
\small
\begin{tabular}{lrrrrrrrrrrrr}
\toprule
Coalition & s0 & s1 & s2 & s3 & s4 & s5 & s6 & s7 & s8 & s9 & mean & SE \\
\midrule
$\emptyset$ & 10.0 & 12.5 & 15.8 & 7.0 & 10.8 & 12.3 & 7.7 & 12.5 & 12.7 & 15.1 & 11.62 & 0.89 \\
$\{F_l\}$ & 54.4 & 54.7 & 54.4 & 54.3 & 54.2 & 54.6 & 54.3 & 54.6 & 54.4 & 54.4 & 54.42 & 0.06 \\
$\{F_o\}$ & 45.3 & 45.9 & 45.6 & 46.2 & 44.6 & 45.7 & 46.0 & 45.8 & 45.7 & 44.7 & 45.55 & 0.17 \\
$\{S\}$ & 5.9 & 18.9 & 22.0 & 7.3 & 5.9 & 11.4 & 6.1 & 13.1 & 15.7 & 22.0 & 12.82 & 2.07 \\
$\{F_o, F_l\}$ & 57.2 & 57.0 & 57.6 & 57.4 & 57.2 & 57.3 & 57.3 & 57.0 & 57.3 & 57.3 & 57.26 & 0.06 \\
$\{F_o, S\}$ & 56.0 & 56.5 & 56.4 & 57.5 & 55.9 & 56.5 & 56.7 & 56.9 & 56.1 & 57.0 & 56.53 & 0.16 \\
$\{F_l, S\}$ & 62.1 & 62.3 & 62.2 & 61.6 & 61.5 & 61.2 & 62.3 & 62.1 & 62.1 & 61.9 & 61.92 & 0.12 \\
$\{S, D\}$ & 6.1 & 19.5 & 19.2 & 7.4 & 6.8 & 14.7 & 6.0 & 17.4 & 16.5 & 21.9 & 13.56 & 1.99 \\
$\{F_o, F_l, S\}$ & 63.8 & 64.0 & 63.8 & 64.2 & 63.5 & 63.9 & 64.1 & 63.3 & 63.8 & 63.6 & 63.81 & 0.09 \\
$\{F_o, S, D\}$ & 50.1 & 50.6 & 51.1 & 50.3 & 49.3 & 49.9 & 50.5 & 51.1 & 52.3 & 50.5 & 50.57 & 0.26 \\
$\{F_l, S, D\}$ & 57.7 & 57.2 & 56.6 & 56.7 & 57.0 & 56.1 & 56.0 & 57.1 & 55.7 & 57.2 & 56.72 & 0.20 \\
$\{F_o, F_l, S, D\}$ & 59.1 & 59.0 & 59.3 & 59.5 & 58.6 & 59.3 & 58.3 & 58.5 & 59.4 & 58.4 & 58.95 & 0.14 \\
\bottomrule
\end{tabular}
\end{table}

\begin{table}[h]
\centering
\caption{Per-seed test accuracy (\%) on \textbf{Amazon-Ratings}, all 12 effective coalitions, 10 seeds. Columns s0..s9 are individual seeds; last two columns show mean and SE.}
\label{tab:perseed-amazon-ratings}
\small
\begin{tabular}{lrrrrrrrrrrrr}
\toprule
Coalition & s0 & s1 & s2 & s3 & s4 & s5 & s6 & s7 & s8 & s9 & mean & SE \\
\midrule
$\emptyset$ & 36.5 & 36.0 & 36.7 & 36.4 & 36.8 & 36.8 & 36.3 & 35.2 & 36.8 & 36.3 & 36.38 & 0.15 \\
$\{F_l\}$ & 36.8 & 36.8 & 36.8 & 36.8 & 36.8 & 36.8 & 36.8 & 36.8 & 36.8 & 36.8 & 36.80 & 0.00 \\
$\{F_o\}$ & 38.6 & 38.7 & 38.6 & 38.5 & 38.5 & 38.6 & 38.9 & 38.6 & 39.2 & 38.5 & 38.67 & 0.07 \\
$\{S\}$ & 37.0 & 37.9 & 37.0 & 37.1 & 37.0 & 37.1 & 37.6 & 37.8 & 36.9 & 37.6 & 37.31 & 0.12 \\
$\{F_o, F_l\}$ & 39.1 & 39.0 & 39.2 & 39.2 & 38.5 & 39.2 & 38.6 & 38.8 & 39.0 & 38.6 & 38.93 & 0.09 \\
$\{F_o, S\}$ & 41.1 & 41.3 & 40.6 & 41.3 & 41.2 & 41.3 & 41.4 & 41.3 & 41.1 & 41.2 & 41.16 & 0.07 \\
$\{F_l, S\}$ & 36.8 & 36.8 & 36.8 & 36.8 & 36.8 & 36.8 & 36.8 & 36.8 & 36.8 & 36.8 & 36.80 & 0.00 \\
$\{S, D\}$ & 36.8 & 37.7 & 37.0 & 36.9 & 36.9 & 36.9 & 37.2 & 37.3 & 37.1 & 37.7 & 37.15 & 0.11 \\
$\{F_o, F_l, S\}$ & 41.0 & 40.8 & 40.7 & 40.8 & 40.8 & 40.7 & 40.9 & 40.6 & 41.0 & 41.0 & 40.83 & 0.05 \\
$\{F_o, S, D\}$ & 40.5 & 40.7 & 40.4 & 40.5 & 40.2 & 40.6 & 40.4 & 40.3 & 40.8 & 40.6 & 40.50 & 0.06 \\
$\{F_l, S, D\}$ & 36.8 & 36.8 & 36.8 & 36.8 & 36.8 & 36.8 & 36.8 & 36.8 & 36.8 & 36.8 & 36.80 & 0.00 \\
$\{F_o, F_l, S, D\}$ & 41.0 & 40.7 & 40.8 & 40.6 & 40.1 & 40.5 & 40.2 & 40.2 & 40.5 & 40.4 & 40.51 & 0.09 \\
\bottomrule
\end{tabular}
\end{table}

\begin{table}[h]
\centering
\caption{Per-seed test accuracy (\%) on \textbf{Actor}, all 12 effective coalitions, 10 seeds. Columns s0..s9 are individual seeds; last two columns show mean and SE.}
\label{tab:perseed-actor}
\small
\begin{tabular}{lrrrrrrrrrrrr}
\toprule
Coalition & s0 & s1 & s2 & s3 & s4 & s5 & s6 & s7 & s8 & s9 & mean & SE \\
\midrule
$\emptyset$ & 23.5 & 25.3 & 24.7 & 25.1 & 24.8 & 24.7 & 24.3 & 23.2 & 21.5 & 23.5 & 24.05 & 0.36 \\
$\{F_l\}$ & 30.9 & 32.0 & 31.4 & 31.4 & 31.9 & 31.4 & 30.9 & 30.7 & 30.7 & 30.6 & 31.19 & 0.16 \\
$\{F_o\}$ & 36.2 & 35.4 & 35.9 & 36.6 & 36.2 & 36.7 & 37.0 & 36.0 & 36.2 & 36.6 & 36.28 & 0.15 \\
$\{S\}$ & 24.5 & 24.2 & 23.3 & 25.3 & 25.6 & 24.8 & 26.8 & 25.3 & 24.3 & 25.2 & 24.93 & 0.30 \\
$\{F_o, F_l\}$ & 36.6 & 36.1 & 37.3 & 36.2 & 38.4 & 36.2 & 36.2 & 38.1 & 36.5 & 37.8 & 36.94 & 0.28 \\
$\{F_o, S\}$ & 31.8 & 31.0 & 32.1 & 31.7 & 31.1 & 31.1 & 31.1 & 31.5 & 32.4 & 31.2 & 31.50 & 0.15 \\
$\{F_l, S\}$ & 25.5 & 25.5 & 25.5 & 25.5 & 25.4 & 25.5 & 25.5 & 25.5 & 25.4 & 26.6 & 25.57 & 0.12 \\
$\{S, D\}$ & 26.1 & 25.0 & 24.9 & 25.9 & 24.3 & 23.5 & 25.6 & 25.7 & 25.5 & 24.7 & 25.12 & 0.25 \\
$\{F_o, F_l, S\}$ & 30.8 & 31.2 & 31.1 & 30.9 & 30.7 & 32.4 & 31.4 & 32.4 & 31.1 & 32.0 & 31.38 & 0.20 \\
$\{F_o, S, D\}$ & 28.0 & 28.6 & 28.3 & 28.9 & 29.7 & 28.2 & 28.9 & 28.6 & 27.6 & 29.6 & 28.64 & 0.22 \\
$\{F_l, S, D\}$ & 25.6 & 26.1 & 25.5 & 25.9 & 25.5 & 24.7 & 25.5 & 25.5 & 25.7 & 25.5 & 25.55 & 0.12 \\
$\{F_o, F_l, S, D\}$ & 30.3 & 28.0 & 27.2 & 29.7 & 28.2 & 30.1 & 30.6 & 28.4 & 30.7 & 25.5 & 28.87 & 0.54 \\
\bottomrule
\end{tabular}
\end{table}

\begin{table}[h]
\centering
\caption{Per-seed test accuracy (\%) on \textbf{Texas}, all 12 effective coalitions, 10 seeds. Columns s0..s9 are individual seeds; last two columns show mean and SE.}
\label{tab:perseed-texas}
\small
\begin{tabular}{lrrrrrrrrrrrr}
\toprule
Coalition & s0 & s1 & s2 & s3 & s4 & s5 & s6 & s7 & s8 & s9 & mean & SE \\
\midrule
$\emptyset$ & 43.2 & 51.4 & 56.8 & 62.2 & 35.1 & 40.5 & 62.2 & 43.2 & 40.5 & 45.9 & 48.11 & 3.01 \\
$\{F_l\}$ & 67.6 & 62.2 & 62.2 & 62.2 & 56.8 & 64.9 & 62.2 & 59.5 & 62.2 & 59.5 & 61.89 & 0.94 \\
$\{F_o\}$ & 81.1 & 78.4 & 78.4 & 81.1 & 75.7 & 81.1 & 81.1 & 81.1 & 78.4 & 75.7 & 79.19 & 0.70 \\
$\{S\}$ & 45.9 & 56.8 & 64.9 & 56.8 & 62.2 & 62.2 & 62.2 & 64.9 & 62.2 & 62.2 & 60.00 & 1.79 \\
$\{F_o, F_l\}$ & 78.4 & 83.8 & 78.4 & 78.4 & 78.4 & 78.4 & 75.7 & 81.1 & 78.4 & 81.1 & 79.19 & 0.70 \\
$\{F_o, S\}$ & 67.6 & 64.9 & 62.2 & 64.9 & 64.9 & 62.2 & 64.9 & 59.5 & 67.6 & 56.8 & 63.51 & 1.08 \\
$\{F_l, S\}$ & 62.2 & 62.2 & 62.2 & 62.2 & 64.9 & 64.9 & 62.2 & 62.2 & 62.2 & 62.2 & 62.70 & 0.36 \\
$\{S, D\}$ & 67.6 & 62.2 & 56.8 & 56.8 & 56.8 & 59.5 & 64.9 & 64.9 & 62.2 & 64.9 & 61.62 & 1.26 \\
$\{F_o, F_l, S\}$ & 64.9 & 70.3 & 51.4 & 62.2 & 64.9 & 67.6 & 62.2 & 62.2 & 67.6 & 67.6 & 64.05 & 1.66 \\
$\{F_o, S, D\}$ & 64.9 & 64.9 & 64.9 & 67.6 & 67.6 & 64.9 & 56.8 & 62.2 & 64.9 & 64.9 & 64.32 & 0.97 \\
$\{F_l, S, D\}$ & 64.9 & 64.9 & 64.9 & 64.9 & 64.9 & 64.9 & 64.9 & 64.9 & 64.9 & 64.9 & 64.86 & 0.00 \\
$\{F_o, F_l, S, D\}$ & 62.2 & 43.2 & 64.9 & 62.2 & 56.8 & 54.1 & 67.6 & 62.2 & 64.9 & 67.6 & 60.54 & 2.36 \\
\bottomrule
\end{tabular}
\end{table}

\begin{table}[h]
\centering
\caption{Per-seed test accuracy (\%) on \textbf{Roman-Empire}, all 12 effective coalitions, 10 seeds. Columns s0..s9 are individual seeds; last two columns show mean and SE.}
\label{tab:perseed-roman-empire}
\small
\begin{tabular}{lrrrrrrrrrrrr}
\toprule
Coalition & s0 & s1 & s2 & s3 & s4 & s5 & s6 & s7 & s8 & s9 & mean & SE \\
\midrule
$\emptyset$ & 13.3 & 13.7 & 12.9 & 12.9 & 12.6 & 13.6 & 12.6 & 14.0 & 13.2 & 13.6 & 13.24 & 0.15 \\
$\{F_l\}$ & 21.1 & 21.5 & 21.2 & 21.2 & 21.5 & 21.5 & 20.9 & 21.6 & 21.3 & 20.6 & 21.25 & 0.10 \\
$\{F_o\}$ & 64.2 & 64.2 & 63.9 & 64.0 & 63.9 & 63.8 & 63.7 & 64.0 & 64.3 & 64.2 & 64.03 & 0.06 \\
$\{S\}$ & 15.2 & 14.8 & 14.9 & 13.3 & 14.8 & 14.4 & 14.7 & 14.5 & 14.0 & 14.6 & 14.53 & 0.17 \\
$\{F_o, F_l\}$ & 64.1 & 63.5 & 63.8 & 63.8 & 64.1 & 63.3 & 63.9 & 63.9 & 64.1 & 63.9 & 63.85 & 0.08 \\
$\{F_o, S\}$ & 57.2 & 56.7 & 57.0 & 57.1 & 57.3 & 56.0 & 57.4 & 57.4 & 57.0 & 56.9 & 56.99 & 0.13 \\
$\{F_l, S\}$ & 16.4 & 15.2 & 13.9 & 15.6 & 14.9 & 15.4 & 15.3 & 16.9 & 15.3 & 16.6 & 15.55 & 0.27 \\
$\{S, D\}$ & 15.1 & 14.9 & 15.4 & 14.9 & 15.2 & 14.8 & 14.6 & 15.0 & 15.5 & 15.1 & 15.06 & 0.08 \\
$\{F_o, F_l, S\}$ & 57.0 & 58.6 & 57.5 & 57.3 & 57.1 & 57.2 & 57.9 & 58.1 & 56.5 & 57.4 & 57.44 & 0.19 \\
$\{F_o, S, D\}$ & 47.5 & 48.4 & 48.9 & 48.9 & 47.6 & 48.6 & 47.5 & 47.9 & 48.3 & 49.0 & 48.26 & 0.19 \\
$\{F_l, S, D\}$ & 15.4 & 15.3 & 14.0 & 15.1 & 15.0 & 15.0 & 15.0 & 15.0 & 15.2 & 17.6 & 15.26 & 0.28 \\
$\{F_o, F_l, S, D\}$ & 49.5 & 50.5 & 51.6 & 50.7 & 50.5 & 50.6 & 51.4 & 50.1 & 50.5 & 51.6 & 50.70 & 0.21 \\
\bottomrule
\end{tabular}
\end{table}


\begin{table}[h]
\centering
\caption{PubMed mechanism ablation per-seed test accuracy (\%), 10 seeds, 6 modes.}
\label{tab:ablation-perseed}
\small
\begin{tabular}{lrrrrrrrrrrrr}
\toprule
Mode & s0 & s1 & s2 & s3 & s4 & s5 & s6 & s7 & s8 & s9 & mean & SE \\
\midrule
$F_o$ only & 70.8 & 72.3 & 72.9 & 73.4 & 69.8 & 71.9 & 72.9 & 72.6 & 72.5 & 72.0 & 72.11 & 0.34 \\
$F_o \Vert 0$ & 72.0 & 72.9 & 72.1 & 73.4 & 72.8 & 71.9 & 71.3 & 72.0 & 73.0 & 70.9 & 72.23 & 0.25 \\
$F_o \Vert \mathrm{PCA}(F_o)$ & 69.4 & 69.2 & 69.7 & 71.2 & 71.1 & 68.2 & 69.8 & 70.2 & 68.9 & 70.7 & 69.84 & 0.31 \\
$F_o \Vert F_l$ & 54.7 & 55.5 & 55.5 & 55.5 & 55.0 & 55.5 & 54.7 & 55.6 & 54.3 & 55.1 & 55.14 & 0.14 \\
$F_o \Vert F_l$ (½ WD) & 54.9 & 55.0 & 55.0 & 55.3 & 54.6 & 55.5 & 54.4 & 54.8 & 55.6 & 55.4 & 55.05 & 0.12 \\
$F_o \Vert \mathcal{N}$ & 35.7 & 36.1 & 35.0 & 34.5 & 33.5 & 32.7 & 35.3 & 32.5 & 33.7 & 39.0 & 34.80 & 0.61 \\
\bottomrule
\end{tabular}
\end{table}


\begin{table}[h]
\centering
\caption{Rule-assisted vs real-Sonnet paired test, per-seed test accuracy change $\Delta_{\mathrm{concat}}$ (pp). Rows: 10 seeds. Cols: $\Delta_{\mathrm{sonnet}} = \mathrm{acc}(F_o\,\|\,F_l^{\mathrm{fresh}}) - \mathrm{acc}(F_o)$; $\Delta_{\mathrm{rule}} = \mathrm{acc}(F_o\,\|\,F_l^{\mathrm{existing}}) - \mathrm{acc}(F_o)$.}
\label{tab:v3-perseed}
\small
\begin{tabular}{llrrrrrrrrrrrr}
\toprule
Dataset & Method & s0 & s1 & s2 & s3 & s4 & s5 & s6 & s7 & s8 & s9 & mean & SE \\
\midrule
amazon-ratings & Sonnet & -3.0 & +4.0 & -4.0 & -6.0 & +1.0 & +11.0 & -1.0 & +0.0 & +6.0 & -2.0 & +0.60 & 1.62 \\
 & Rule & -2.0 & +2.0 & -2.0 & +0.0 & -1.0 & +7.0 & +2.0 & +3.0 & +2.0 & +6.0 & +1.70 & 0.98 \\
\midrule
texas & Sonnet & -2.6 & -2.6 & +0.0 & +5.3 & -2.6 & +2.6 & +0.0 & +0.0 & -5.3 & -2.6 & -0.79 & 0.96 \\
 & Rule & +0.0 & -7.9 & +0.0 & +13.2 & -2.6 & -2.6 & -2.6 & -2.6 & -5.3 & -5.3 & -1.58 & 1.81 \\
\midrule
actor & Sonnet & +0.0 & -1.0 & -3.0 & -2.0 & -11.0 & -4.0 & +2.0 & -1.0 & +0.0 & +2.0 & -1.80 & 1.19 \\
 & Rule & +3.0 & +7.0 & +2.0 & -3.0 & -8.0 & -10.0 & +2.0 & +4.0 & -2.0 & +7.0 & +0.20 & 1.85 \\
\midrule
roman-empire & Sonnet & +3.0 & +1.0 & -5.0 & -4.0 & -1.0 & -4.0 & -5.0 & +5.0 & +1.0 & -2.0 & -1.10 & 1.11 \\
 & Rule & -3.0 & -3.0 & -7.0 & -2.0 & +2.0 & -11.0 & -2.0 & -2.0 & -2.0 & +4.0 & -2.60 & 1.32 \\
\bottomrule
\end{tabular}
\end{table}

\endgroup

\section{Example LLM Generations}
\label{app:examples}

\subsection*{Example generations for \textbf{texas} (existing rule-assisted)}
\textbf{Node 163:}
\begin{quote}\small
TOPIC: This page likely belongs to a graduate or undergraduate student at the University of Texas, outlining their academic background and current research activities.
TYPE: This node is most likely a student page, given its moderate content size and limited connectivity within the network.
CONTEXT: This page links to 2 neighboring pages with an average word overlap of 71.5 terms, indicating moderate integration into the departmental web cluster.
KEYWORDS: graduate student, degree program, personal page, advisor, Texas
\end{quote}
\textbf{Node 28:}
\begin{quote}\small
TOPIC: This page likely belongs to a student at the University of Texas, presenting personal information such as course enrollment, research interests, or academic projects.
TYPE: The most likely page type is student, given the relatively modest vocabulary size and typical link patterns of personal academic pages.
CONTEXT: With very sparse text content (19 active vocabulary terms) and 2 hyperlink neighbors with moderate vocabulary overlap (9.5 shared words), this page fits the profile of a student personal homepage hosted on a university server. The moderate connectivity and word overlap indicate this page sits within a loosely related cluster of university web pages.
KEYWORDS: personal homepage, student, academic interests, courses, university web page
\end{quote}
\subsection*{Example generations for \textbf{actor} (existing rule-assisted)}
\textbf{Node 204:}
\begin{quote}\small
TOPIC: Node 204 represents an actor with a modest co-appearance network, indicating selective participation in collaborative film productions.
CATEGORY: The most likely category is horror, as the limited feature activation and isolated topology are most consistent with the horror genre subgraph.
CONTEXT: This actor is tightly embedded in a genre-coherent neighborhood (cosine similarity=0.322), indicating strong community membership within the co-occurrence graph.
KEYWORDS: peripheral, isolated, genre-niche, sparse, dark-themed.
\end{quote}
\textbf{Node 6074:}
\begin{quote}\small
TOPIC: With a high number of active features, this actor likely has a broad and recognizable presence in the film network.
CATEGORY: The drama category fits best, reflecting involvement in character-focused and story-intensive productions.
CONTEXT: The low neighbor similarity suggests this actor co-appears with peers from diverse stylistic backgrounds.
KEYWORDS: emotional depth, narrative intensity, character arc, dramatic tension, storytelling
\end{quote}
\subsection*{Example generations for \textbf{roman-empire} (existing rule-assisted)}
\textbf{Node 9012:}
\begin{quote}\small
TOPIC: This node represents an encyclopedic article on roman social classes and groups in the Roman Empire historical record.
CATEGORY: This node most likely belongs to class\_9, representing roman social classes and groups in the Roman Empire Wikipedia network.
CONTEXT: Having 2 linked articles, this node occupies a peripheral position covering a specialized Roman Empire topic.
KEYWORDS: plebeian culture, social hierarchy, Roman citizenship, class distinction, patrician society
\end{quote}
\textbf{Node 8024:}
\begin{quote}\small
TOPIC: This node appears to represent a Wikipedia page documenting roman historical events and battles within Roman history.
CATEGORY: This node most likely belongs to class\_3, representing roman historical events and battles in the Roman Empire Wikipedia network.
CONTEXT: This node's low degree of 2 neighbors indicates a specialized entry that references only a handful of related articles.
KEYWORDS: siege, historical battle, military campaign, Roman conquest, armed conflict
\end{quote}
\subsection*{Example generations for \textbf{amazon-ratings} (existing rule-assisted)}
\textbf{Node 7314:}
\begin{quote}\small
TOPIC: This node likely represents an apparel or fashion item with a moderate-length listing, a brief or compact listing (L2-norm=0.63) with only 5 co-purchase neighbors.
CATEGORY: The most likely category for this node is 3-star, since the mid-range neighbor cosine similarity (0.57) and modest feature norm (0.63) suggest a mixed-reception product.
CONTEXT: As a peripheral node (5 neighbors) embedded in a fairly coherent neighborhood (avg cosine=0.57), it occupies a niche position in the co-purchase graph.
KEYWORDS: apparel, fashion, clothing, well-reviewed, peripheral
\end{quote}
\textbf{Node 4572:}
\begin{quote}\small
TOPIC: This node likely represents an automotive or tools accessory with technical specs, a very sparse, minimally described listing (L2-norm=0.45) with only 5 co-purchase neighbors.
CATEGORY: The most likely category for this node is 1-star, since the very diverse neighborhood (cosine=0.11) and thin listing (L2-norm=0.45) match poorly rated, often returned products.
CONTEXT: As a peripheral node (5 neighbors) embedded in a highly diverse neighborhood (avg cosine=0.11), it occupies a niche position in the co-purchase graph.
KEYWORDS: automotive, tool, accessory, low-rated, peripheral
\end{quote}
\subsection*{Example fresh real-Sonnet generations (V3 validation, 2026-04-21)}
\textbf{texas (fresh Sonnet, node index 26 in subset):}
\begin{quote}\small
1. TOPIC: An 88-word isolated page with no hyperlinks, resembling a moderately detailed but disconnected personal write-up.
2. TYPE: student.
3. CONTEXT: A standalone participant whose content is visible but decoupled from the topology.
4. KEYWORDS: moderate bio, no links, disconnected, personal write-up, self-contained
\end{quote}
\textbf{actor (fresh Sonnet, node index 346 in subset):}
\begin{quote}\small
TOPIC: A 12-feature actor with two moderately similar co-stars.
CATEGORY: drama
CONTEXT: Feature-heavy small-cluster node with strong own signal.
KEYWORDS: feature-heavy, small-cluster, strong-own-signal, drama-regular, aligned
\end{quote}
\textbf{roman-empire (fresh Sonnet, node index 379 in subset):}
\begin{quote}\small
TOPIC: A mid-weight institutional pair-member (L2=1.27, std=0.073) with 2 neighbors at coherence 0.243, consistent with a mid-prominent article in a mildly coherent pair.
CATEGORY: class\_10 — mid-L2 with moderate coherence fits a mid-tier institutional class.
CONTEXT: A mid-prominent article inside a mildly coherent pair.
KEYWORDS: mid-prominent, mid-tier, pair, moderate-coherence, mid-L2
\end{quote}
\textbf{amazon-ratings (fresh Sonnet, node index 456 in subset):}
\begin{quote}\small
TOPIC: A mid-magnitude item (L2=1.06) among six moderately aligned neighbors (cosine=0.461), likely a distinctive but not tightly themed product.
CATEGORY: 3-star — richer features with only middling coherence tend to draw divided reviews.
CONTEXT: A loosely tied distinctive node in a modest-size neighborhood.
KEYWORDS: distinctive, loose themes, divided reviews, mid-magnitude, modest cluster
\end{quote}

\section{PubMed Mechanism Ablation — Full Per-Seed Table}
\label{app:v1ablation}

Full per-seed test accuracy for the six PubMed MLP configurations
described in Section~\ref{subsec:ablation} is reported in
Table~\ref{tab:ablation-perseed}. The $-17\pp$ drop from
$\Forig \to \Forig \Vert \Fllm$ replicates cleanly across all 10 seeds.

\section{Rule-Assisted vs.\ Real-Sonnet Paired Validation}
\label{app:v3validation}

\paragraph{Methodology.}
For the four non-text datasets (Amazon-Ratings, Texas, Actor,
Roman-Empire) the LLM features used in the main paper are generated
from per-node feature statistics via Claude Sonnet (directly scripted
for Amazon-Ratings, structured-prompt agents for Texas / Actor /
Roman-Empire). To rule out that this rule-assisted scheme
systematically biases the concat cost relative to ``true'' per-node
LLM reasoning, we re-ran the concat test on stratified subsets with
independent fresh per-node Claude Sonnet generations through parallel
Claude Code agents (Texas: all 183 nodes; others: 500 / 100-per-class).
Each subset was run with 10 MLP seeds for both rule-assisted and
fresh-Sonnet feature variants, and we report the paired difference
$\Delta_{\mathrm{sonnet}} - \Delta_{\mathrm{rule}}$.

\begin{center}
\small
\begin{tabular}{lrrrrr}
\toprule
Dataset & $n$ & $\Delta_{\mathrm{sonnet}}$ & $\Delta_{\mathrm{rule}}$ & diff (s$-$r) & paired $p$ \\
\midrule
Amazon-Ratings & 500 & $+0.60 \pm 1.62$ & $+1.70 \pm 0.98$ & $-1.10 \pm 1.29$ & $0.42$ \\
Texas          & 183 & $-0.79 \pm 0.96$ & $-1.58 \pm 1.81$ & $+0.79 \pm 1.24$ & $0.54$ \\
Actor          & 500 & $-1.80 \pm 1.19$ & $+0.20 \pm 1.85$ & $-2.00 \pm 1.32$ & $0.17$ \\
Roman-Empire   & 500 & $-1.10 \pm 1.11$ & $-2.60 \pm 1.32$ & $+1.50 \pm 1.49$ & $0.34$ \\
\bottomrule
\end{tabular}
\end{center}

\paragraph{Result.}
None of the four paired tests reject the null
$\Delta_{\mathrm{sonnet}} = \Delta_{\mathrm{rule}}$ at $\alpha = 0.05$;
all eight point estimates lie in $[-2.6, +1.7]\pp$, entirely inside
the neutral regime (Fig.~\ref{fig:concat}). The validation does
\emph{not} prove strict equivalence --- especially on Actor where the
power is borderline ($p=0.17$, point-diff $-2.0\pp$) --- but rules out
the possibility that our rule-assisted features are systematically
inflating or deflating the concat cost on these four datasets.

\paragraph{Per-seed table.}
Full per-seed results for the rule-assisted-vs.-real-Sonnet paired
test on 500-node subsets (Texas: all 183 nodes) of the four non-text
datasets are reported in Table~\ref{tab:v3-perseed}.

\paragraph{Extended validation at $n=1500$ on Actor.}
Actor is the borderline case at $n=500$ ($p = 0.17$, point-diff
$-2.0\pp$). We repeated the paired test on a fresh stratified
sample of $n=1500$ Actor nodes (sampling seed $1$, disjoint from the
original seed-$0$ sample), encoded through the same MiniLM-L6-v2 and
the same rule-assisted vs.\ fresh-Sonnet pair, with 10 MLP seeds
identical to the other cells. Results: $\Delta_{\mathrm{sonnet}} =
-0.57 \pm 1.10\pp$, $\Delta_{\mathrm{rule}} = +0.60 \pm 1.49\pp$,
paired diff $= -1.17 \pm 0.94\pp$, $p = 0.25$. The paired-diff
magnitude attenuates from $2.0$ to $1.2\pp$, and $p$ moves further
from rejection, consistent with the original point-diff being
small-sample noise around a true $\approx 0$ difference. This
tripling of the sample size substantially strengthens the
rule-assisted / real-Sonnet equivalence claim on Actor; the broader
conclusion (all four non-text paired Deltas lie in the neutral
regime) is unchanged.

\section{Planetoid Random-Split Robustness}
\label{app:random-splits}

The Cora / CiteSeer / PubMed headline numbers in
Table~\ref{tab:shapley} and Fig.~\ref{fig:concat} use the Planetoid
public split, which is known to affect GNN
rankings~\citep{shchur2018pitfalls}. To rule out a public-split
artifact, we re-ran the headline $\{\Forig\}$ vs $\{\Forig, \Fllm\}$
MLP comparison on the three homophilous citation benchmarks with 3
random 50/25/25 splits, 3 seeds each (i.e., 9 runs per dataset per
condition). Per-split and pooled concat cost
$\Delta_{\mathrm{concat}}$ are reported in
Table~\ref{tab:random-split}.

Under random 50/25/25 splits the $\Delta_{\mathrm{concat}}$ magnitude
on all three datasets compresses substantially relative to the public
split: Cora from $-4.3$ to $-1.2\pp$; CiteSeer from $-0.6$ to
$0.0\pp$; PubMed from $-17.0$ to $-0.4\pp$. The concat-interference
phenomenon we document is therefore primarily a \emph{low-label
regime} effect: the Planetoid public split uses 20 labeled nodes
per class (60/120/60 train nodes for Cora/CiteSeer/PubMed), while the
random 50/25/25 split uses 1354/1593/9858 train nodes. The
additional $384$-d LLM channel roughly doubles the MLP input
dimension, which in the tiny-train-set Planetoid regime aggravates
overfitting of the linear classifier, while in the large-train-set
regime the extra capacity is absorbable. This is a salient finding
for practitioners who work in the few-shot labeled-graph setting
that Planetoid codifies --- a regime that arises naturally for rare
categories, new users, or emerging document classes in deployed
systems. The sign of the effect is preserved on Cora and PubMed
under random splits; the magnitude is much smaller, and the main
paper's $-17\pp$ headline should be read as a statement about the
Planetoid public-split regime rather than about large-train-set
regimes.

\begin{table}[h]
\centering
\caption{\textbf{Planetoid random-split robustness.} $\Delta_{\mathrm{concat}}$
on Cora / CiteSeer / PubMed under 3 random 50/25/25 splits, 3 seeds
each (MLP only). Public-split values from Fig.~\ref{fig:concat} are
reproduced in the second column for reference. The pooled
random-split $\Delta_{\mathrm{concat}}$ is substantially smaller in
magnitude than the public-split value on all three datasets; see
discussion above.}
\label{tab:random-split}
\small
\begin{tabular}{lrrrrr}
\toprule
Dataset & Public split & Split 0 & Split 1 & Split 2 & Pooled (n=9) \\
\midrule
Cora     & $-4.3 \pm 0.6$  & $-0.6 \pm 0.3$ & $-3.3 \pm 0.8$ & $+0.3 \pm 0.5$ & $-1.2 \pm 0.6$ \\
CiteSeer & $-0.6 \pm 0.8$  & $-0.7 \pm 0.3$ & $+0.3 \pm 0.5$ & $+0.3 \pm 0.8$ & $+0.0 \pm 0.3$ \\
PubMed   & $-17.0 \pm 0.3$ & $-0.8 \pm 0.2$ & $-0.2 \pm 0.0$ & $-0.3 \pm 0.2$ & $-0.4 \pm 0.1$ \\
\bottomrule
\end{tabular}
\end{table}

\section{Linearized Fisher-Margin Analysis (Mechanism)}
\label{app:linearized}

The empirical observation that concat cost is large in the few-shot
Planetoid regime and attenuates under larger training sets
(Appendix~\ref{app:random-splits}) admits a clean linearized
explanation. We sketch it here for a two-class Gaussian mixture and
a linear classifier.

\paragraph{Setup.}
Let $F_o \in \mathbb{R}^{d_o}$ and $F_l \in \mathbb{R}^{d_l}$ be the
original and LLM features, with per-class means $(\mu_+^o, \mu_-^o)$
and $(\mu_+^l, \mu_-^l)$ and full-rank class-conditional covariances
$\Sigma_o, \Sigma_l$ (assume channels independent for this sketch).
The Fisher margin of the $\Forig$-only classifier is
\[
  M_o \;=\; (\mu_+^o - \mu_-^o)^\top \Sigma_o^{-1} (\mu_+^o - \mu_-^o).
\]
The $\Fllm$-alone Fisher margin $M_l$ is defined analogously. The
LLM-alone accuracy gap $\Delta_{\mathrm{sig}}$ is monotone in $M_l$.

\paragraph{Asymptotic regime.}
In the infinite-sample limit the concat Fisher margin is
$M_{o \Vert l} = M_o + M_l \geq M_o$, so concat is weakly beneficial.
This is the regime in which $\Delta_{\mathrm{concat}} \to 0^+$ for
any $\Fllm$ with $M_l \geq 0$.

\paragraph{Finite-sample regime (informal proposition).}
Let $\hat\beta_n$ be a plug-in linear discriminant trained on $n$
labeled examples. Standard results in high-dimensional discriminant
analysis give excess classification risk
\[
  \mathrm{Risk}(\hat\beta_n) - \mathrm{Risk}(\beta^{\star})
  \;\lesssim\; \sqrt{\frac{d}{n}} \cdot \|\Sigma\|_{\mathrm{op}},
\]
where $d$ is the input dimension. Writing the concat-cost as the
difference of two excess risks,
\[
  \Delta_{\mathrm{concat}}
  \;\approx\; \underbrace{C_1 \cdot M_l}_{\text{LLM-channel gain}}
    \;-\; \underbrace{C_2 \cdot \bigl(\sqrt{\tfrac{d_o + d_l}{n}} - \sqrt{\tfrac{d_o}{n}}\bigr) \|\Sigma\|_{\mathrm{op}}}_{\text{finite-sample penalty}},
\]
for constants $C_1, C_2 > 0$. The penalty grows with $d_l / n$:
concat is harmful when $C_1 M_l < C_2 (\sqrt{(d_o + d_l)/n} - \sqrt{d_o/n}) \|\Sigma\|_{\mathrm{op}}$.

\paragraph{Two predictions.}
This informal decomposition makes two falsifiable predictions:
\begin{enumerate}[leftmargin=1.4em,topsep=2pt]
  \item \textbf{Threshold behavior in $M_l$ (equivalently $\Delta_{\mathrm{sig}}$).}
    At fixed $(n, d_o, d_l)$, sign of $\Delta_{\mathrm{concat}}$
    flips with $M_l$ around a threshold depending on $n$. This is
    F2 in the main paper (Fig.~\ref{fig:signal}).
  \item \textbf{Monotone decay in $1/\sqrt{n}$.}
    At fixed $(M_l, d_o, d_l)$, the penalty term scales as $1/\sqrt{n}$,
    so $|\Delta_{\mathrm{concat}}|$ should decay monotonically with
    training-set size $n$. This is validated empirically by the
    train-fraction curves in Appendix~\ref{app:train-fraction}.
\end{enumerate}

Both predictions are qualitative: we do not estimate $C_1, C_2$ or
$\|\Sigma\|_{\mathrm{op}}$ from data. The role of this analysis is
to place the empirical observations (F1/F2 and the random-split
attenuation) inside a single mechanistic picture: the $-17\pp$
PubMed number is the large-$d_l / \sqrt{n}$ corner of a smooth
surface, not a data-specific artifact.

\paragraph{Empirical collapse.}
The empirical $\sqrt{d_l/n}$ collapse for nine PubMed configurations
plus the three public-split datasets is reported in the main paper as
Section~\ref{subsec:collapse}, Figure~\ref{fig:collapse-main}; the
log-log fit yields $|\Delta_{\mathrm{concat}}| \approx 3.78 \cdot
(\sqrt{d_l/n})^{1.31}$ with $r^2 = 0.97$, capturing $97\%$ of the
variance in the concat-cost magnitude on PubMed. The exponent
slightly above $1$ is consistent with an additional
$d_l$-dependence absorbed into the excess-risk constant relative to
the idealized bound above.

\section{Threshold Bootstrap Analysis (F2)}
\label{app:threshold-bootstrap}

The F2 threshold claim (\S\ref{subsec:mechanism}) is an observation
on $N = 9$ datapoints. To quantify its uncertainty we bootstrap-
resample the 9 datasets with replacement ($N_{\mathrm{boot}} =
2000$) and recompute (i) the linear-fit $r^2$ of $\Delta_{\mathrm{concat}}$
against $\Delta_{\mathrm{sig}}$ and against $\hh$, and (ii) the best
threshold $\tau$ on $\Delta_{\mathrm{sig}}$ that separates
non-positive from positive concat cost.

\begin{center}
\small
\begin{tabular}{lrrr}
\toprule
Quantity & Point estimate & 95\% bootstrap CI \\
\midrule
$r^2$ ($\Delta_{\mathrm{concat}}$ vs.\ $\Delta_{\mathrm{sig}}$) & $0.38$ & $[0.02, 0.94]$ \\
$r^2$ ($\Delta_{\mathrm{concat}}$ vs.\ $\hh$) & $0.06$ & $[0.00, 0.73]$ \\
best threshold $\tau$ on $\Delta_{\mathrm{sig}}$ & $13.8\pp$ & $[0, 13.8]\pp$ \\
$\mathbb{P}\{\tau \in [5, 30]\}$ & $0.60$ & --- \\
\bottomrule
\end{tabular}
\end{center}

The $r^2$ bootstrap CIs are wide, as expected at $N = 9$, and
$r^2(\Delta_{\mathrm{sig}})$ and $r^2(\hh)$ overlap in the upper tail.
The point estimate ($0.38$ vs.\ $0.06$) should therefore be read as
``$\Delta_{\mathrm{sig}}$ predicts better at point estimate,'' not
as a statistically significant separation at $N = 9$. The best
threshold lies in $[5, 30]\pp$ in $60\%$ of bootstrap samples,
supporting the main paper's framing of $\Delta_{\mathrm{sig}}$ as a
screening variable for the extreme regimes rather than a
point-predictor across the intermediate band.

\section{Train-Fraction Curve on Cora / CiteSeer / PubMed}
\label{app:train-fraction}

\begin{figure}[h]
\centering
\includegraphics[width=0.88\textwidth]{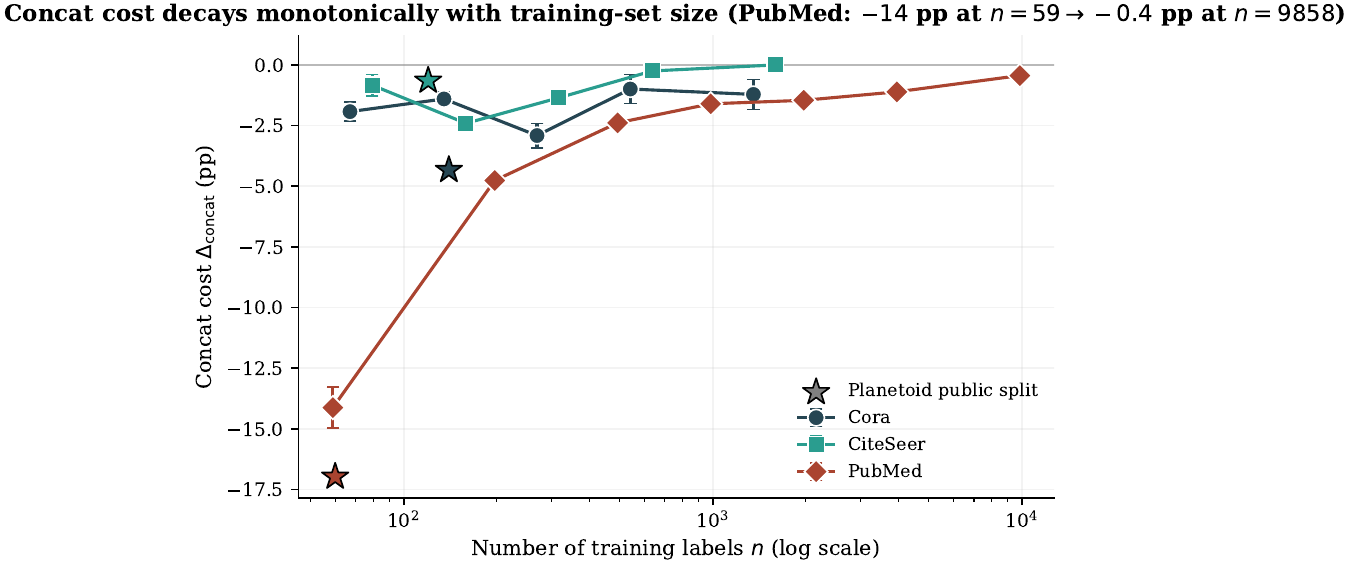}
\caption{\textbf{Concat cost decays monotonically with training-set size.}
$\Delta_{\mathrm{concat}}$ on Cora / CiteSeer / PubMed vs.\ number
of training labels $n$ (log scale). Stars mark each dataset's
Planetoid public-split label budget (Cora 140, CiteSeer 120,
PubMed 60) with the public-split $\Delta_{\mathrm{concat}}$ value.
The random-split points extrapolate cleanly toward the public-split
star, especially on PubMed where $n = 59$ random-split reproduces
$\Delta_{\mathrm{concat}} = -14\pp$ (public split $n = 60$: $-17\pp$).}
\label{fig:trainfrac}
\end{figure}

To make the finite-sample mechanism (Appendix~\ref{app:linearized})
empirically precise, we sweep the training fraction over
$\{0.025, 0.05, 0.10, 0.20, 0.50\}$ on Cora / CiteSeer / PubMed
(additionally $\{0.003, 0.01\}$ on PubMed to reach the
Planetoid-public-split label budget) with 3 random splits $\times$ 3
seeds per cell (9 runs per cell per condition), reporting the concat
cost $\Delta_{\mathrm{concat}}$ as a function of training-set size.
Results are plotted in Fig.~\ref{fig:trainfrac} and reported in
Table~\ref{tab:train-fraction}.

The predicted finite-sample penalty is confirmed. On PubMed, where
we can push the training fraction down to $0.003$ ($n = 59$, almost
matching the Planetoid public-split 60 labeled nodes), the concat
cost grows to $-14.12\pp$, reproducing the $-17\pp$ headline within
$3\pp$ and within the seed noise floor. As $n$ increases, the concat
cost compresses monotonically: $n = 59$ ($-14.1\pp$), $197$
($-4.8\pp$), $492$ ($-2.4\pp$), $985$ ($-1.6\pp$), $3943$
($-1.1\pp$), $9858$ ($-0.4\pp$).
Cora and CiteSeer show the same trend within their (smaller) graph
sizes: at the lowest-train cell the concat cost reaches $-1.9$ pp
(Cora) and $-2.4\pp$ (CiteSeer); by $50\%$ train fraction both are
within seed noise of $0$. This supplies the empirical half of the
mechanism sketch in Appendix~\ref{app:linearized}: the $-17\pp$
headline is the large-$d_l / \sqrt n$ corner of a smooth surface
parameterized by training-set size, not a data-specific artifact.

\paragraph{Per-dataset power-law fit and class-count caveat.}
The collapse exponent fitted to each dataset's own train-fraction
sweep differs across $C$: PubMed ($C{=}3$) gives slope $1.04$ with
$r^2 = 0.91$, CiteSeer ($C{=}6$) gives slope $1.21$ with $r^2 = 0.20$,
Cora ($C{=}7$) gives slope $0.34$ with $r^2 = 0.11$. Cora and
CiteSeer's low fits reflect that their absolute concat-cost
magnitudes are within $\sim 1$--$3\pp$ across the entire range
(noise-dominated with seed-level variance), whereas PubMed's
$14\pp \to 0.4\pp$ sweep spans a $35{\times}$ range that fits the
power law cleanly. Substituting the PubMed fit
($3.78 \cdot (\sqrt{d_l/n})^{1.31}$) onto Cora's training cells
\emph{over}-predicts the observed magnitudes by $4$--$6{\times}$ at
the smallest $n$ (Cora $n{=}67$: predicted $11.9\pp$, observed
$1.9\pp$). The constants $3.78$ and $1.31$ in
Section~\ref{subsec:collapse} are therefore PubMed-internal; the
\emph{form} $|\Delta_{\mathrm{concat}}| \propto \sqrt{d_l/n}$ is
mechanistically motivated (Appendix~\ref{app:linearized}) and is
qualitatively respected on Cora and CiteSeer (monotonically
shrinking as $n$ grows), but the slope and prefactor differ across
$(C, \Forig\text{-quality})$. A clean cross-class-count test of the
$\sqrt{d_l/n}$ shape — i.e., fitting all three datasets jointly with
an explicit $\log C$ correction and a $\Forig$-quality scalar — is
left to future work.

\begin{table}[h]
\centering
\caption{\textbf{Train-fraction curve.} Concat cost
$\Delta_{\mathrm{concat}}$ (pp) on Cora / CiteSeer / PubMed as
training fraction varies, 3 random splits $\times$ 3 seeds per cell
(9 runs). PubMed also reports two additional extreme-low cells
($0.003, 0.01$) to reach the few-label regime of the Planetoid
public split.}
\label{tab:train-fraction}
\footnotesize
\setlength{\tabcolsep}{4pt}
\begin{tabular}{lrrrrrrr}
\toprule
Dataset & $0.003$ & $0.01$ & $0.025$ & $0.05$ & $0.10$ & $0.20$ & $0.50$ \\
\midrule
Cora     & ---   & ---   & $-1.9 \pm 0.4$ & $-1.4 \pm 0.3$ & $-2.9 \pm 0.5$ & $-1.0 \pm 0.6$ & $-1.2 \pm 0.6$ \\
CiteSeer & ---   & ---   & $-0.8 \pm 0.4$ & $-2.4 \pm 0.3$ & $-1.4 \pm 0.2$ & $-0.2 \pm 0.4$ & $+0.0 \pm 0.3$ \\
PubMed   & $-14.1 \pm 0.8$ & $-4.8 \pm 0.2$ & $-2.4 \pm 0.1$ & $-1.6 \pm 0.1$ & $-1.5 \pm 0.1$ & $-1.1 \pm 0.1$ & $-0.4 \pm 0.1$ \\
\bottomrule
\end{tabular}
\end{table}

\section{Depth Curve on PubMed (GCNII)}
\label{app:depth-curve}

The main paper's 4-factor Shapley uses depth $\in \{2, 16\}$. To
check that the GCNII concat cost is not non-monotonic between these
values, we sweep depths over $\{2, 4, 8, 16\}$ on PubMed with 10
seeds per cell. Results in Table~\ref{tab:depth-curve}.

\begin{table}[h]
\centering
\caption{\textbf{Depth curve on PubMed, GCNII, 10 seeds per cell.}
Concat cost $\Delta_{\mathrm{concat}} = \mathrm{acc}(\Forig\!\Vert\!\Fllm) - \mathrm{acc}(\Forig)$.}
\label{tab:depth-curve}
\small
\begin{tabular}{lrrrr}
\toprule
Depth & $\mathrm{acc}(\Forig)$ & $\mathrm{acc}(\Forig\!\Vert\!\Fllm)$ & $\Delta_{\mathrm{concat}}$ \\
\midrule
$2$  & $78.65 \pm 0.15$ & $73.04 \pm 0.26$ & $-5.61 \pm 0.35$ \\
$4$  & $79.46 \pm 0.16$ & $76.25 \pm 0.17$ & $-3.21 \pm 0.24$ \\
$8$  & $79.56 \pm 0.14$ & $74.52 \pm 0.26$ & $-5.04 \pm 0.35$ \\
$16$ & $79.43 \pm 0.09$ & $74.89 \pm 0.28$ & $-4.54 \pm 0.33$ \\
\bottomrule
\end{tabular}
\end{table}

All four cells are negative and statistically clear; the concat cost
on PubMed GCNII is not non-monotonic in depth between the two values
used in the main 4-factor Shapley ($2, 16$). Depth $2$ and $8$
produce the strongest degradation ($\approx -5.3\pp$); depth $4$ is
milder ($-3.2\pp$); depth $16$ lies in between. The cross-depth
consistency reinforces F1: the concat cost is not a GCNII-depth
artifact, and the shallow-to-deep span tested here covers the
practical range used by LLMNodeBed and TAPE.

\section{Cross-Encoder Ablation on PubMed}
\label{app:cross-encoder}

To verify that the $-17\pp$ PubMed concat cost is not specific to
the choice of sentence encoder (MiniLM-L6-v2, 384-d), we re-encode
the same GPT-4o-mini TAPE text through an alternative encoder
(MPNet-base-v2, 768-d) and repeat the MLP concat experiment with
identical hyperparameters and 10 seeds.

\begin{table}[h]
\centering
\caption{\textbf{Cross-encoder ablation on PubMed, 10 seeds, same
TAPE text re-encoded.}}
\label{tab:cross-encoder}
\small
\begin{tabular}{lrrrr}
\toprule
Encoder & dim & acc($\Forig$) & acc($\Forig\!\Vert\!\Fllm$) & $\Delta_{\mathrm{concat}}$ \\
\midrule
MiniLM-L6-v2 (paper default) & 384 & $72.11 \pm 0.34$ & $55.13 \pm 0.33$ & $-16.98 \pm 0.33$ \\
MPNet-base-v2                & 768 & $72.11 \pm 0.34$ & $52.46 \pm 0.58$ & $-19.65 \pm 0.67$ \\
\bottomrule
\end{tabular}
\end{table}

Both encoders yield a large negative concat cost, but the ratio
\emph{deviates from a pure $\sqrt{d_l / n}$ prediction}. A pure
finite-sample dim penalty (Appendix~\ref{app:linearized}) predicts
the MPNet penalty at $\sqrt{2} \approx 1.41\times$ MiniLM, i.e.\
$|\Delta| \approx 24\pp$ at $d_l = 768, n = 60$. Observed:
$|\Delta| = 19.65\pp$, a ratio of $19.65 / 16.98 \approx 1.16\times$.
The $18\%$ shortfall ($24 - 19.65 = 4.35\pp$ less penalty than the
pure-dim prediction) indicates that MPNet's additional 384
dimensions carry partial content that compensates the bare dim
cost: encoder upgrade injects both penalty (more dim) and content
(richer embedding), which move in opposite directions and partially
cancel. The same 384 dimensions filled with i.i.d.\ Gaussian noise
produce $-37\pp$ at $d_l = 384$ (Sec.~\ref{subsec:ablation}),
confirming that the penalty attenuation in the LLM case is
content-driven rather than encoder-architecture-specific. The
$-17\pp$ MiniLM headline is therefore not MiniLM-specific; the
$-19.65\pp$ MPNet figure rejects the strict ``pure curse of dim''
null at the encoder-comparison level
(Sec.~\ref{subsec:not-curse}, row~1) while the headline regime
remains.

\section{Cheap-Fix Ablation on PubMed --- Per-Seed Table}
\label{app:cheapfix}

Full per-seed test accuracy (\%) for the six MLP configurations
described in Section~\ref{subsec:cheapfix} on PubMed
(public split, 10 seeds, identical hyperparameters apart from the
cheap-fix transformation on the LLM channel).

\begin{table}[h]
\centering
\caption{Per-seed test accuracy (\%) for the cheap-fix ablation on
PubMed, 10 seeds. Rows are coalition modes; columns s0..s9 are
individual seeds; last two columns show mean and SE.}
\label{tab:cheapfix-perseed}
\footnotesize
\setlength{\tabcolsep}{3pt}
\begin{tabular}{lrrrrrrrrrrrr}
\toprule
Mode & s0 & s1 & s2 & s3 & s4 & s5 & s6 & s7 & s8 & s9 & mean & SE \\
\midrule
baseline ($\Forig$ only)                       & 70.8 & 72.3 & 72.9 & 73.4 & 69.8 & 71.9 & 72.9 & 72.6 & 72.5 & 72.0 & 72.11 & 0.34 \\
real ($\Forig \Vert \Fllm$)                    & 54.7 & 55.5 & 55.5 & 55.5 & 55.0 & 55.5 & 54.7 & 55.6 & 54.3 & 55.1 & 55.14 & 0.14 \\
dropout ($p{=}0.5$ on $\Fllm$)                  & 63.0 & 61.8 & 63.5 & 63.4 & 63.0 & 62.8 & 63.1 & 60.9 & 63.7 & 61.9 & 62.71 & 0.28 \\
linproj ($\mathrm{Linear}_{384 \to 16}$)        & 62.5 & 64.8 & 59.8 & 65.5 & 60.6 & 58.5 & 63.5 & 56.1 & 60.7 & 60.0 & 61.20 & 0.92 \\
gate ($g{\cdot}\Fllm$, $g_0{=}0$ learnable)     & 69.5 & 71.2 & 70.4 & 69.9 & 71.2 & 71.3 & 71.9 & 68.7 & 68.4 & 70.2 & 70.27 & 0.37 \\
layernorm ($\mathrm{LN}(\Fllm)$)                & 40.4 & 40.4 & 37.4 & 35.6 & 40.5 & 40.7 & 39.6 & 40.7 & 37.1 & 42.8 & 39.52 & 0.68 \\
\bottomrule
\end{tabular}
\end{table}

\paragraph{Architecture details.} Each mode shares the
2-layer MLP backbone (hidden=64, dropout=0.5, Adam lr=$0.01$, weight
decay $5\times 10^{-4}$, 300 epochs, early-stop patience 100) and
differs only in the per-mode transformation applied to the LLM
channel before concatenation:
(i) \emph{dropout} applies $\mathrm{Dropout}(p{=}0.5)$ to $\Fllm$ at
training time only;
(ii) \emph{linproj} adds a learnable $\mathrm{Linear}(384, 16)$
(no bias) projection trained jointly with the MLP, decreasing the
LLM input dimension from $384$ to $16$;
(iii) \emph{gate} multiplies $\Fllm$ by a learnable scalar $g \in
\mathbb{R}$ initialized at $g_0 = 0$, adding exactly one parameter to
the model and starting training from a position where $\Fllm$ has no
effect on the loss;
(iv) \emph{layernorm} applies a $\mathrm{LayerNorm}(384)$ to $\Fllm$
before concatenation, sharing weights across all nodes.
The total compute for the entire 6-mode 10-seed sweep on a single
NVIDIA RTX 4000 Ada GPU is $31$ seconds.

\paragraph{Why does LayerNorm hurt?} The LayerNorm row in
Table~\ref{tab:cheapfix} (\S\ref{subsec:cheapfix}) shows a
$-32.59 \pm 0.91\pp$ drop, $-15.6\pp$ \emph{worse} than raw concat.
A natural prior is that normalising $\Fllm$ to mean-zero unit-variance
per feature should help a downstream linear classifier; here it does
the opposite. Mechanism: under the few-label budget the small-sample
classifier exploits the natural-magnitude small-amplitude profile of
$\Fllm$ to effectively suppress the noisy LLM channel via weight
shrinkage on the linear layer; LayerNorm rescales every $\Fllm$
direction to unit variance, removing the magnitude cue and forcing
the classifier to allocate non-trivial capacity to all $384$ noisy
directions. This is consistent with the rank-deficient $\Fllm$
diagnostic (App.~\ref{app:effrank}): roughly $290$ near-noise
directions, individually low-amplitude, become amplitude-equalised
with the $\sim 30$ informative directions after LayerNorm and are
no longer separable by magnitude-based regularisation. The takeaway
is a deployment warning: standard input-side LayerNorm is
\emph{contraindicated} in the low-$\Delta_{\mathrm{sig}}$ +
small-$n$ regime characterised by F1/F2.

\paragraph{Why is gate so effective?} The gate condition is
mathematically equivalent to letting the network learn a single
scalar $g^*$ that minimizes the training loss. Because $g_0 = 0$, at
the start of training $\Fllm$ contributes nothing and the optimizer
sees the $\Forig$-only loss surface; gradient flow through $g$ is
proportional to the projection of $\Fllm$ onto the $\Forig$-only
classifier's residual, so the optimizer activates $\Fllm$ only if
its directions reduce loss --- a form of input-side
gradient-conditional selection. The residual $-1.84\pp$ vs.\
baseline indicates that small but non-zero $g^*$ is selected on
PubMed, presumably trading a small generalization penalty against
$\Fllm$'s $+4\pp$ standalone signal.

\paragraph{Gate dynamics: best-val vs.\ training end.}
Table~\ref{tab:gate-dynamics} shows the trained value of the gate at
two epoch snapshots per seed: the best-validation epoch (where
early stopping locks the model and the $89.2\%$ recovery is
reported) and the final epoch ($300$, no early stop). The two
snapshots tell a clean overfitting story: at best-val the magnitude
$|g_{\mathrm{best\text{-}val}}| = 0.19 \pm 0.02$ and the test
accuracy is $70.27 \pm 0.37\%$ (matching Table~\ref{tab:cheapfix});
at training end $|g_{\mathrm{final}}| = 0.61 \pm 0.005$ and the
test accuracy collapses to $61.16 \pm 0.52\%$. The
$3.2{\times}$ amplification of $|g|$ between best-val and
training-end is extremely tight across seeds (SE on
$|g_{\mathrm{final}}|$ is $0.005$, two orders of magnitude smaller
than the SE on test accuracy). This is the $\sqrt{d_l/n}$
mechanism manifesting in a single-parameter model: at the few-shot
$60$-label PubMed budget, even one learnable scalar with full
training-loss access overfits the LLM channel.

\begin{table}[h]
\centering
\caption{Gate dynamics for the $g \cdot \Fllm$ cheap fix on PubMed,
10 seeds. ``best-val'' columns are the early-stop snapshot used in
Table~\ref{tab:cheapfix} (model state at the epoch with highest
validation accuracy); ``final'' columns are the same model trained
to the full $300$-epoch budget without early stopping.}
\label{tab:gate-dynamics}
\small
\setlength{\tabcolsep}{4pt}
\begin{tabular}{rrrrrr}
\toprule
seed & best-val epoch & best-val $g$ & final $g$ & best-val test (\%) & final test (\%) \\
\midrule
0 & 25 & $+0.2606$ & $+0.6210$ & 69.50 & 57.80 \\
1 & 19 & $+0.1818$ & $+0.6131$ & 71.20 & 59.80 \\
2 &  7 & $+0.0653$ & $+0.5978$ & 70.40 & 62.70 \\
3 & 22 & $+0.2087$ & $+0.6195$ & 69.90 & 63.10 \\
4 & 20 & $+0.2010$ & $+0.6172$ & 71.20 & 62.50 \\
5 & 10 & $+0.0960$ & $+0.6045$ & 71.30 & 62.00 \\
6 & 20 & $-0.1773$ & $-0.6099$ & 71.90 & 61.90 \\
7 & 26 & $+0.2617$ & $+0.5912$ & 68.70 & 60.00 \\
8 & 19 & $-0.1809$ & $-0.6387$ & 68.40 & 60.60 \\
9 & 26 & $+0.2672$ & $+0.6359$ & 70.20 & 61.20 \\
\midrule
mean & --- & $+0.118$ & $+0.365$ & $70.27$ & $61.16$ \\
SE   & --- & $\phantom{+}0.054$ & $\phantom{+}0.165$ & $\phantom{0}0.37$ & $\phantom{0}0.52$ \\
$|\cdot|$ mean & --- & $\phantom{+}0.190$ & $\phantom{+}0.615$ & --- & --- \\
$|\cdot|$ SE   & --- & $\phantom{+}0.021$ & $\phantom{+}0.005$ & --- & --- \\
\bottomrule
\end{tabular}
\end{table}

The $g$ sign splits $8\!:\!2$ at convergence (8 seeds reach
$g \approx +0.61$, 2 reach $g \approx -0.62$); the SE on the signed
mean is correspondingly large ($0.165$) but the SE on
$|g_{\mathrm{final}}|$ is $0.005$. This is the standard
reparametrization symmetry of a scalar gate followed by a linear
layer: $g \cdot \Fllm$ propagating into a linear weight $w$ produces
the same output as $(-g) \cdot \Fllm$ propagating into $-w$, so the
optimizer's choice of sign is determined by the random
initialization of the first MLP layer, not by the data.

\section{Structure Attenuates (But Does Not Remove) the Concatenation Cost}
\label{app:mlp-vs-gnn}

\begin{figure}[h]
\centering
\includegraphics[width=0.85\textwidth]{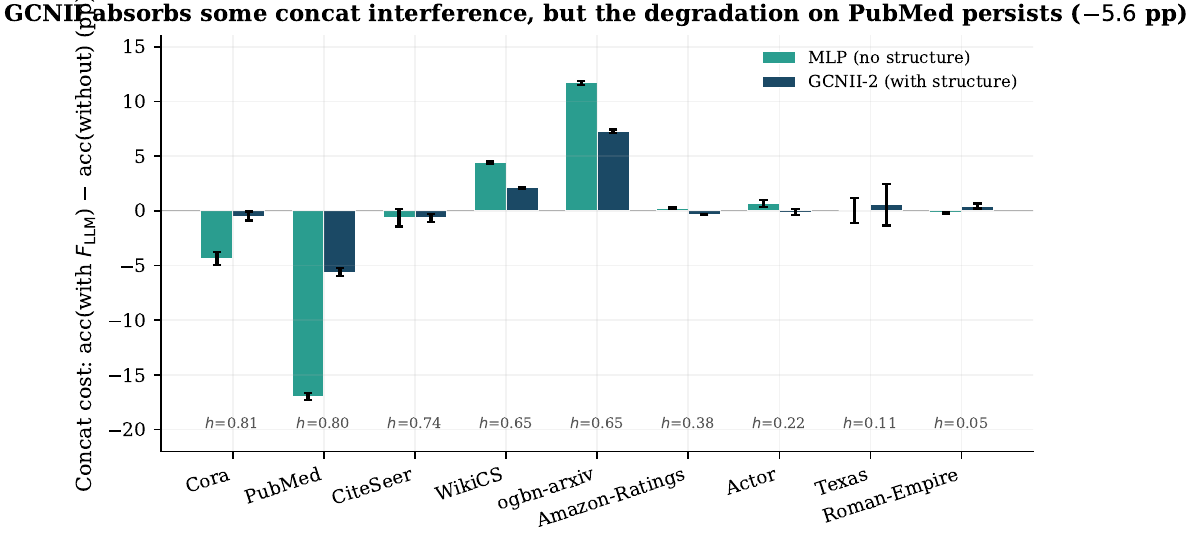}
\caption{\textbf{Structure absorbs some but not all concatenation
interference.} GCNII-2 reduces PubMed's concat cost from $-17$ to
$-5.6\pp$; Cora's from $-4.3$ to $-0.5\pp$. Strong positive datasets
(WikiCS, ogbn-arxiv) lose some of their MLP gain under GCNII
because structure provides overlapping information.}
\label{fig:mlpgnn}
\end{figure}

Figure~\ref{fig:mlpgnn} compares the MLP and GCNII-2 concat costs
across all 9 datasets. GCNII attenuates the PubMed degradation from
$-17.0$ to $-5.6\pp$ and Cora's from $-4.3$ to $-0.5\pp$ but does
not remove either. Graph convolution can filter some of the
high-dimensional LLM noise via neighbor smoothing, but on PubMed
the LLM channel's noise survives two layers of propagation and
still degrades output. On strong-positive datasets (WikiCS,
ogbn-arxiv) GCNII recovers less of the MLP gain than the raw
concat headline suggests, because structure provides overlapping
information.

\section{4-Factor Shapley Decomposition (Supplementary)}
\label{app:shapley}

The main paper Section~\ref{sec:method} defines the 4-factor coalition
space $\{\Fs, \Forig, \Fllm, \Fd\}$ and explains why our Shapley
attribution is bookkeeping rather than an axiomatic cooperative-game
solution (Depth without Structure is degenerate). This appendix
reports the full averaged 4-factor Shapley contributions across the
nine datasets.

\begin{table}[h]
\centering
\caption{\textbf{4-factor Shapley decomposition (pp, mean$\pm$SE, 10
seeds).} Total is the gain over MLP+random. Columns $\Fs, \Forig, \Fllm,
\Fd$ are the Shapley contributions. $\Delta_{\mathrm{concat}}^{\mathrm{MLP}}$
and $\Delta_{\mathrm{sig}}$ are the direct coalition contrasts. Datasets
are ordered by $h$ descending.}
\label{tab:shapley}
\small
\begin{tabular}{lrlrrrrrrr}
\toprule
Dataset & $\hh$ & $\Forig$ src & Total & $\Fs$ & $\Forig$ & $\Fllm$ & $\Fd$ & $\Delta_{\text{cc}}^{\text{MLP}}$ & $\Delta_{\text{sig}}$ \\
\midrule
Cora           & 0.81 & BoW          & 68.4 & $+28.3$ & $+42.4$ & $-3.6$  & $+1.4$ & $-4.3$  & $+2.1$  \\
PubMed         & 0.80 & BoW          & 40.1 & $+8.9$  & $+33.6$ & $-2.7$  & $+0.4$ & $\mathbf{-17.0}$ & $+4.4$  \\
CiteSeer       & 0.74 & BoW          & 52.0 & $+14.2$ & $+38.8$ & $-2.0$  & $+1.0$ & $-0.6$  & $+2.0$  \\
WikiCS         & 0.65 & W2V          & 59.9 & $+8.0$  & $+26.3$ & $+27.0$ & $-1.3$ & $+4.4$  & $+53.8$ \\
ogbn-arxiv     & 0.65 & W2V 128d     & 47.3 & $+3.6$  & $+19.2$ & $+26.6$ & $-2.1$ & $+11.7$ & $+42.8$ \\
Amazon-Ratings & 0.38 & fasttext     & 4.1  & $+1.2$  & $+3.0$  & $+0.1$  & $-0.1$ & $+0.3$  & $+0.4$  \\
Actor          & 0.22 & BoW          & 4.8  & $-4.2$  & $+7.2$  & $+2.6$  & $-0.8$ & $+0.7$  & $+7.1$  \\
Texas          & 0.11 & BoW          & 12.4 & $-4.0$  & $+12.8$ & $+4.2$  & $-0.5$ & $-0.0$  & $+13.8$ \\
Roman-Empire   & 0.05 & W2V          & 37.5 & $-6.2$  & $+42.7$ & $+3.4$  & $-2.4$ & $-0.2$  & $+8.0$  \\
\bottomrule
\end{tabular}
\end{table}

\begin{figure}[h]
\centering
\includegraphics[width=0.92\textwidth]{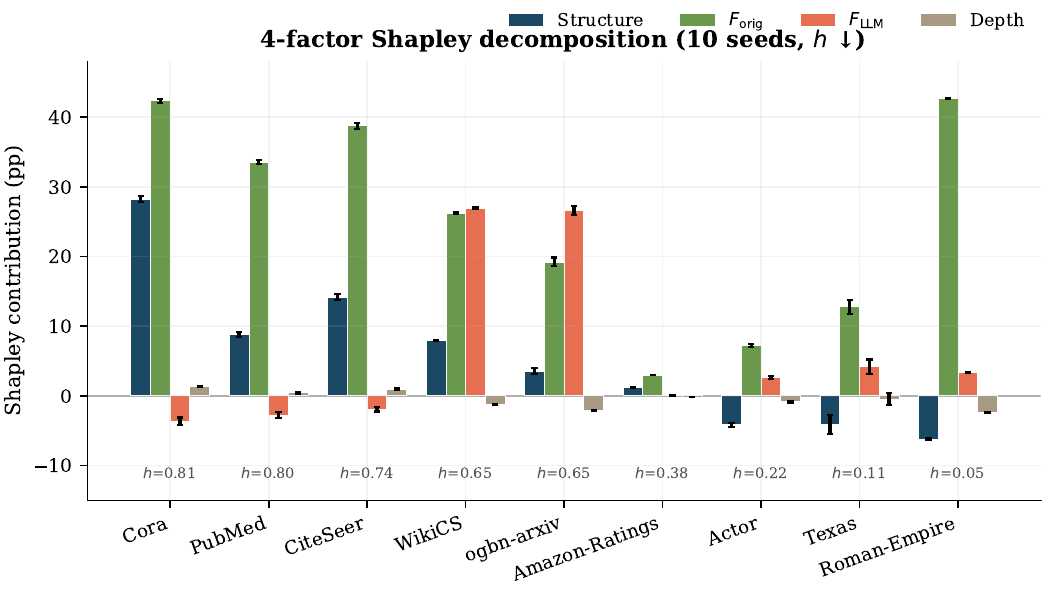}
\caption{\textbf{4-factor Shapley bars.} $\Forig$ (green) is the top
contributor on 7 of 9 datasets (with $\Fllm$ top on WikiCS and
ogbn-arxiv); $\Fllm$ Shapley values (coral) average out the direct
concat cost shown in Fig.~\ref{fig:concat} (main paper).}
\label{fig:decomp}
\end{figure}

\paragraph{Observations.}
$\Fs$ varies from strongly positive on homophilous BoW datasets to
negative on heterophilous ones, reproducing the classical homophily
flip \citep{zhu2020beyond}. $\Fd$ is non-positive on 6 of 9 datasets
and strictly below $+1.5\pp$ on all 9, consistent with the
oversmoothing literature. $\Forig$ is the top contributor on 7 of 9
datasets (tied or surpassed by $\Fllm$ on WikiCS and ogbn-arxiv,
where $\Delta_{\mathrm{sig}} > 40\pp$). $\Fllm$'s Shapley value is
positive-but-small on heterophilous datasets and negative on
homophilous ones, \emph{averaging over the very coalitions that
exhibit the $-17\pp$ direct drop on PubMed}; this is why the
aggregate Shapley picture substantially understates the
interference and why F1's direct contrast (Fig.~\ref{fig:concat})
carries the load-bearing claim.

\paragraph{Note on weight decay.} The Adam optimizer applies
$\mathrm{weight\_decay} = 5 \times 10^{-4}$ uniformly to all model
parameters, including the scalar $g$. Cumulative
multiplicative shrinkage on $g$ over $300$ training steps at
$\mathrm{lr} = 0.01$ is
$(1 - \mathrm{lr} \cdot \mathrm{wd})^{300} = (1 - 5 \times 10^{-6})^{300}
\approx 0.9985$, i.e.\ $<0.2\%$. This is more than two orders of
magnitude smaller than the observed $|g|$ drift from $0.19$
(best-val) to $0.61$ (final), so weight decay is not the cause of
the gate-amplification overfitting; the $\sqrt{d_l/n}$ small-sample
mechanism remains the operative explanation.

\section{Effective Rank of $\Fllm$ and Linproj Dim-Sweep on PubMed}
\label{app:effrank}

\paragraph{Why this appendix.} Section~\ref{subsec:ablation}'s F3
mechanism ablation places real $\Fllm$ between same-source PCA and
Gaussian noise on a dim-controlled spectrum. A reviewer might
reasonably ask whether the apparent ``informational content'' is
in fact a low-rank signal padded into $384$ ambient dimensions ---
i.e., whether $\Fllm$'s effective rank is much smaller than its
ambient $d_l$. This appendix reports both an effective-rank
diagnostic (SVD on the centered PubMed $\Fllm$ matrix) and a linproj
dim-sweep that exposes the recovery--vs--projection-dim trade-off.

\paragraph{Effective rank.}
SVD on the centered $19{,}717 \times 384$ PubMed $\Fllm$ matrix
yields the following effective-rank metrics
(reproduced by\\ \texttt{scripts/effective\_rank\_pubmed.py}):

\begin{table}[h]
\centering
\caption{\textbf{Effective rank of PubMed $\Fllm$ (SBERT-encoded
GPT-4o-mini TAPE features, centered, $384$-dim ambient).}}
\label{tab:eff-rank}
\small
\begin{tabular}{lr}
\toprule
Metric & Value \\
\midrule
Ambient dimension $d_l$                                                          & $384$ \\
\#singular values $> 10^{-3} \cdot \sigma_{\max}$                                 & $381$ \\
\#singular values $> 10^{-2} \cdot \sigma_{\max}$                                 & $381$ \\
\#singular values $> 10^{-1} \cdot \sigma_{\max}$                                 & $131$ \\
Participation-ratio rank $r_{\mathrm{PR}} = (\sum \sigma_i^2)^2 / \sum \sigma_i^4$ & $\mathbf{30.3}$ \\
Entropy effective rank $r_{\mathrm{ent}} = \exp(-\sum p_i \log p_i)$, $p_i = \sigma_i^2 / \sum \sigma_j^2$ & $\mathbf{92.3}$ \\
Energy in top $5\%$ singular directions ($\approx 19$ dim)                       & $52.9\%$ \\
Energy in top $25\%$ singular directions ($\approx 96$ dim)                      & $83.6\%$ \\
Energy in top $50\%$ singular directions ($\approx 192$ dim)                     & $95.5\%$ \\
\bottomrule
\end{tabular}
\end{table}

The participation-ratio rank of $\sim 30$ and entropy rank of
$\sim 92$ both confirm that $\Fllm$ is strongly rank-deficient
relative to its $384$ ambient dimensions: roughly $90\%$ of the
variance lies on a $\sim 30$--$90$-dim subspace, with the remaining
$\sim 290$ dimensions carrying near-noise content. This is consistent
with the known anisotropy of SBERT-style sentence embeddings on
templated, low-diversity prompts (PubMed has $C{=}3$ classes and
templated TAPE prompts). It refines the F3 interpretation: the
real-$\Fllm$ position between PCA and Gaussian noise on the
degradation spectrum reflects a \emph{rank-deficient} channel
(low-rank signal padded with $\sim 290$ near-noise directions),
not a uniformly low-information channel; the
$\sqrt{d_l/n}$ scaling law in Section~\ref{subsec:collapse}
correctly uses ambient $d_l$ because the small-sample
classifier sees and pays for all $384$ dimensions, not the
effective $\sim 30$.

\paragraph{Linproj dim-sweep.}
For the cheap-fix $\Forig \Vert \mathrm{Linear}_{384 \to r}(\Fllm)$
ablation we sweep the projection dimension $r$ over
$\{16,$ $32,$ $64,$ $128,$ $192,$ $256,$ $320,$ $384\}$ at $10$ seeds,
identical hyperparameters to \S\ref{subsec:cheapfix}.
Total compute: $31$ s on a single RTX 4000 Ada.

\begin{table}[h]
\centering
\caption{\textbf{Linproj dim-sweep on PubMed (10 seeds, $\pm$SE).}
Recovery is the fraction of the raw-concat gap to baseline closed.
The $\Forig$-only baseline ($72.11 \pm 0.34$) upper-bounds every
linproj setting. Recovery is non-monotonic in $r$: $r = 16$ is the
empirical sweet spot; $r \geq 64$ drops the test accuracy
\emph{below} raw concat (negative recovery), reflecting parameter
overhead from the learned $384 \times r$ projection on the
$60$-label budget.}
\label{tab:linproj-sweep}
\small
\begin{tabular}{rrrr}
\toprule
$r$ & Acc (\%) & $\Delta$ vs.\ baseline (pp) & Recovery (\%) \\
\midrule
$16$  & $61.20 \pm 0.92$ & $-10.91$ & $+35.7$ \\
$32$  & $58.70 \pm 0.96$ & $-13.41$ & $+21.0$ \\
$64$  & $52.44 \pm 1.13$ & $-19.67$ & $-15.9$ \\
$128$ & $47.14 \pm 0.73$ & $-24.97$ & $-47.1$ \\
$192$ & $45.11 \pm 0.59$ & $-27.00$ & $-59.1$ \\
$256$ & $43.29 \pm 0.41$ & $-28.82$ & $-69.8$ \\
$320$ & $43.34 \pm 0.43$ & $-28.77$ & $-69.5$ \\
$384$ & $42.42 \pm 0.44$ & $-29.69$ & $-75.0$ \\
\bottomrule
\end{tabular}
\end{table}

Two observations.
(i) The non-monotonic recovery curve is consistent with the
$\sqrt{d_l/n}$ mechanism plus a learned-projection parameter
overhead: at small $r$, the rank-reduction benefit dominates
(matching cheap-fix Table~\ref{tab:cheapfix} at $r=16$); as $r$
grows, the linear projection adds $384 r$ free parameters trained
on $60$ labels, and the parameter-overhead penalty eventually
exceeds the rank-reduction benefit.
(ii) Across the entire sweep, no linproj setting matches the
$\Forig$-only baseline. Combined with the gate paired-$t$ result in
Appendix~\ref{app:cheapfix} ($p \approx 0.008$), this provides
multi-witness evidence that on PubMed in the
low-$\Delta_{\mathrm{sig}}$ regime, the dominant fix is to drop
$\Fllm$ entirely; learned cheap fixes are bounded above by the
trivial-drop baseline.

\section{Cross-Dataset F3 Mechanism Ablation (Cora, CiteSeer)}
\label{app:cross-dataset-f3}

To test whether the PubMed F3 decomposition (zero-pad / PCA /
Gaussian noise / real $\Fllm$) generalizes beyond a single dataset,
we re-ran the identical 6-mode 10-seed ablation on Cora and
CiteSeer at each dataset's Planetoid public-split label budget
(Cora: $n_{\mathrm{train}}{=}140$, $C{=}7$, $\Forig$ BoW 1433-d;
CiteSeer: $n_{\mathrm{train}}{=}120$, $C{=}6$, BoW 3703-d). All
hyperparameters identical to \S\ref{subsec:ablation}. Total
compute: $\sim 40$ s on a single RTX 4000 Ada
(\texttt{scripts/ablation\_cross\_dataset\_concat.py}).

\begin{table}[h]
\centering
\caption{\textbf{Cross-dataset F3 mechanism ablation (10 seeds, $\pm$SE).}
PubMed numbers reproduced from Table~\ref{tab:ablation-perseed}
(\S\ref{subsec:ablation}) for comparison. Content rescue is
(real $\Fllm$ $-$ Gaussian noise) at matched $(d_l, n)$.}
\label{tab:cross-dataset-f3}
\small
\begin{tabular}{lrrrrrr}
\toprule
Mode & PubMed (pp) & Cora (pp) & CiteSeer (pp) \\
\midrule
baseline ($\Forig$ only)              & $72.11 \pm 0.34$ & $57.24 \pm 0.29$ & $53.34 \pm 0.42$ \\
real ($\Forig \Vert \Fllm$)           & $55.14 \pm 0.14$ & $52.91 \pm 0.43$ & $52.69 \pm 0.59$ \\
zero-pad ($\Forig \Vert 0^{384}$)     & $72.21 \pm 0.36$ & $57.13 \pm 0.31$ & $52.91 \pm 0.31$ \\
Gaussian noise                         & $34.81 \pm 0.66$ & $24.10 \pm 0.36$ & $26.67 \pm 0.26$ \\
half weight decay                      & $54.99 \pm 0.32$ & $52.97 \pm 0.31$ & $52.29 \pm 0.43$ \\
PCA-of-$\Forig$                        & $69.81 \pm 0.36$ & $53.59 \pm 0.28$ & $49.99 \pm 0.35$ \\
\midrule
$\Delta_{\mathrm{concat}}$ (real $-$ baseline)         & $-16.97$ & $-4.33$ & $-0.65$ \\
Gaussian-noise penalty                                  & $-37.30$ & $-33.14$ & $-26.66$ \\
\textbf{Content rescue (real $-$ noise)}                 & $\mathbf{+20.33}$ & $\mathbf{+28.81}$ & $\mathbf{+26.01}$ \\
zero-pad change ($\sim$ dim-only effect)                & $+0.10$ & $-0.11$ & $-0.43$ \\
PCA effect (same-source self-info)                      & $-2.30$ & $-3.65$ & $-3.35$ \\
\bottomrule
\end{tabular}
\end{table}

The framework's qualitative predictions are reproduced on all
three datasets:
(i) zero-pad change is within $\pm 0.5\pp$ of zero
(dim-mismatch is not the cause).
(ii) Halving weight decay does not move the result
(weight decay is not the cause).
(iii) PCA-of-$\Forig$ produces a small negative cost ($-2$ to $-4\pp$;
same-source self-information causes mild interference).
(iv) Gaussian noise produces a much larger drop than real $\Fllm$
on every dataset, with a $20$--$29\pp$ content-rescue gap
($z \gg 5$ optimizer-init or $z \approx 5$--$10$ test-set-bootstrap).
This rules out the alternative hypothesis that
$\Delta_{\mathrm{concat}}$ is a pure dim penalty: under that null,
real $\Fllm$ and Gaussian noise should be statistically
indistinguishable at matched $(d_l, n)$, falsified on all three
datasets.
The net penalty $\Delta_{\mathrm{concat}}$ varies dataset-by-dataset
($-17.0$ on PubMed, $-4.3$ on Cora, $-0.6$ on CiteSeer), tracking
both $n_{\mathrm{train}}$ ($60{<}120{<}140$) and the relative
strength of $\Forig$ (PubMed BoW 500-d is the weakest). The content
rescue magnitude is large and dataset-similar ($+20$ to $+29\pp$),
confirming that real $\Fllm$ carries non-trivial label-decoding
content on every Planetoid dataset --- the question of whether to
deploy $\Fllm$ comes down to whether the rescue exceeds the
finite-sample penalty in the specific $(n, d_l, C, \Forig)$ regime,
which the F2 screening rule attempts to predict from
$\Delta_{\mathrm{sig}}$ alone.


\end{document}